\documentclass{article}

\PassOptionsToPackage{square,numbers}{natbib}
\usepackage[preprint]{neurips_2023}

\usepackage[utf8]{inputenc} 
\usepackage[T1]{fontenc}    
\usepackage{hyperref}       
\usepackage{url}            
\usepackage{booktabs}       
\usepackage{amsfonts}       
\usepackage{nicefrac}       
\usepackage{microtype}      
\usepackage{xcolor}         
\usepackage{caption}
\usepackage{subcaption}
\usepackage{graphicx}
\usepackage{amsmath}
\usepackage{bm} 
\usepackage{physics} 

\title{\textit{SANE}: The phases of gradient descent through \textbf{S}harpness \textbf{A}djusted \textbf{N}umber of \textbf{E}ffective parameters}

\author{
Lawrence Wang \\
Department of Engineering \\
Oxford University\\
\texttt{lawrencewang@exeter.ox.ac.uk}\\
\And
Stephen J. Roberts \\
Department of Engineering \\
Oxford University\\
\textit{sjrob@robots.ox.ac.uk}\\
}

\begin{document}

\newcommand{\neff}{$N_\mathrm{eff}$}

\newcommand{\lammax}{$\lambda_\mathrm{max}$}
\newcommand{\vmax}{$v_\mathrm{max}$}

\newcommand{\lr}{$\eta$}
\newcommand{\rotins}{$\overline{S(v)}$}
\newcommand{\gins}{$\overline{S(g)}$}
\newcommand{\hessvar}{$\sigma_{\lambda_\mathrm{bulk}}$}
\newcommand{\EF}{\textit{effective dimensionality}}
\newcommand{\dof}{\textit{d.o.f.}}
\newcommand{\dofs}{\textit{d.o.f.}s}

\newcommand{\sane}{SANE}
\newcommand{\eos}{\textit{Edge of Stability}}
\newcommand{\gf}{\textit{smooth}}
\newcommand{\Gf}{\textit{Gradient flow}}
\newcommand{\alf}{$\alpha$}
\newcommand{\fcite}{\textcolor{red}{find cite}}
\newcommand{\redhess}{$\mathcal{H}^\mathrm{r}_k$}

\newcommand{\app}{\textbf{SM}}

\maketitle

\begin{abstract}

Modern neural networks are undeniably successful. Numerous studies have investigated how the curvature of loss landscapes can affect the quality of solutions. In this work we consider the Hessian matrix during network training. We reiterate the connection between the number of ``well-determined'' or ``effective'' parameters and the generalisation performance of neural nets, and we demonstrate its use as a tool for model comparison. By considering the local curvature, we propose \textbf{S}harpness \textbf{A}djusted \textbf{N}umber of \textbf{E}ffective parameters (SANE), a measure of \textit{effective dimensionality} for the quality of solutions. We show that \sane\ is robust to large learning rates, which represent learning regimes that are attractive but (in)famously unstable. We provide evidence and characterise the Hessian shifts across ``loss basins'' at large learning rates. Finally, extending our analysis to deeper neural networks, we provide an approximation to the full-network Hessian, exploiting the natural ordering of neural weights, and use this approximation to provide extensive empirical evidence for our claims.

\end{abstract}

\section{Introduction}\label{section:intro}
With advancements in computation and increased availability of large datasets, deep neural networks have become extremely popular in many machine learning applications ranging from forecasting \cite{Lim_2021}, to vision \cite{simonyan2015vgg,vaswani2017attention}, and language modelling \cite{devlin2019bert,brown2020language}. For each problem, the dataset, network architecture, and objective function jointly define the non-convex landscape through which optimisation algorithms must traverse and locate minimums for network parameters. As we uncover the power of deep neural nets, the \textit{generalisation gap} is revealed to be more pervasive and significantly limits their performance. In this paper, we propose \textbf{Sharpness Adjusted Number of Effective parameters} (\sane) as a measure for model performance and for generalisation, exploiting the intuition that models with fewer effective parameters represent hypotheses that reduce the \textit{generalisation gap}. As we explore the effectiveness of \sane, we uncover insights into the phases of gradient descent. 

Recent studies on generalisation have focused on the taxonomy of minima in the loss landscape \cite{nakkiran2019deep, fort2020deep, li2018visualizing}. Naturally, the study of the second-order characteristics of the \textit{loss landscape} through the Hessian (of the loss w.r.t. network parameters) \cite{Ghorbani2019, Bottcher2022} uncovers geometrical details to describe weight space and allow efficient optimisation steps up to the (local) limits of the quadratic approximation. Numerous theoretical analyses and empirical evidence have suggested that the sharpness of the curvature contributes negatively to generalisation performance \cite{keskar2017, hoffer2018train, jastrzębski2019relation} and \textit{Sharpness Aware Minimisation} \cite{Foret2020} has been proposed as a regularisation objective to leverage this connection. An immediate objection to this approach stems from \textit{Goodhart's law}, which states: "when a measure becomes a target, it ceases to be a good measure" \cite{Strathern1997ImprovingRA}. Additionally, recent works \cite{granziol2020flatness, Kaur2022} have shown that we can manipulate the curvature of landscapes through the intimate relationship it has with the learning rate, to further confound the purported link between curvature and generalisation. Finally, \citet{cohen2022gradient} showed that models can continue to improve their performance despite moving through regions of instabilty, which challenges existing wisdom\cite{lecun1992} for learning rate selection. This encourages the large learning rate regime - trading the non-monotonicity of loss for more effective steps across weight space.

In light of the weakened connection between sharpness and generalisation, \sane\ serves as an alternative to sharpness-based measures for the quality of solutions and for generalisation. \sane\ achieves this by leveraging the connection to ``effective'' eigenvectors of the Hessian, each of which corresponds to a salient direction in weight space that controls significant degrees of freedom as determined by the dataset. In this work, we study gradient descent with large learning rates - a learning regime that allows optimisers to improve the speed of finding good solutions for deep neural nets. We show empirically that \sane\ is a robust measure of model performance under these settings. 

Our contributions are as follows: 
\begin{enumerate}
    \item We introduce \sane, a novel measure of \EF\ for model performance
    \item We show that \sane\ correlates well with generalisation, and we demonstrate its utility as a tool for model selection \textit{post-training} and \textit{during-training}
    \item Through \sane, we uncover insights into the instabilities of gradient descent and study the effects of popular learning rate schedules with these instabilities
    \item We introduce an empirical approximation to the Hessian, with enables the scaling of Hessian computations to deeper neural nets, and we use this approximation to provide extensive empirical justification for our claims on benchmark datasets
    \item To encourage openness and reproducibility of research, we share our code in the \app
\end{enumerate}
We introduce the notation and the motivation for our work in Section \ref{section:background}, and we present our main findings in Section \ref{section:main}. Section \ref{section:related} places our work in the broader literature, and we validate our findings on CIFAR-10 in Section \ref{section:experiments}. We offer a brief discussion and conclude in Section \ref{section:discussion}.

\section{Background}\label{section:background}

\textbf{Notation.} In this work we consider a supervised classification problem where $\bm{x_i} = \{x_i, y_i\}$ constitutes an input-label pair. We parameterise the predictions with a deep neural net with weights $\theta \in \mathbb{R}^{n} = \Theta$ to obtain a prediction function $\hat{y}_i = f_\theta(x_i)$. The loss function is averaged over the training data set $\mathcal{L}_{\theta} = \mathbb{E}_{\bm{x_i}} L_{\theta}(\bm{x_i})$, where $L_{\theta}$ is the cross-entropy loss between the prediction $\hat{y}_i$ and the true label $y_i$. So, we can write the gradient $g = \grad_{\theta} \mathcal{L}_{\theta}$ and the Hessian $\mathcal{H}=\grad^2_{\theta} \mathcal{L}_{\theta}$. We order the eigenvalues of the Hessian in descending order: $\lambda_\mathrm{max}=\lambda_1>\lambda_2>...>\lambda_n=\lambda^{-}_\mathrm{max}$.

In the following sections, we use cosine similarity to measure directional alignment $\mathcal{S}(u) = |\mathcal{S}_c(u_{t}, u_{t+1})|$ and misalignment $\overline{\mathcal{S}(u)} = |1 - \mathcal{S}_c(u_{t}, u_{t+1})|$ where $\mathcal{S}_c$ is the cosine similarity function. 

\textbf{Flat minima.} Flat (wide) regions in weight space are distinguished from sharp (narrow) regions if the objective function changes slowly to shifts in the parameters. This concept has been discussed frequently in the machine learning literature, with a focus on its ability to determine whether models can generalise to unseen data. \citet{Hochreiter1997} provided justification for this connection through the minimum description length framework, suggesting that flat minima permit the greatest compression of data. \citet{djcmThesis} showed, from a Bayesian perspective, that flat minima can be the consequence of an Occam's razor penalty. To estimate local flatness, we adopt the standard approach in the literature which uses a quadratic approximation in the local weight space and equates the top eigenvalue (\lammax) of the Hessian to the sharpness (inverse-flatness) of the loss landscape. 

\textbf{Large learning rates.} Suitable learning rates for gradient descent will naturally differ depending on the factors that influence the weight space, such as the dataset and the neural architecture used. Optimisers, such as ADAM \cite{kingma2017adam}, have been developed for automatic preconditioning, but finding the right learning rates can remain an empirical endeavour in practice. Some learning rates lead to model divergence. For a convex quadratic function $f(x) = \frac{1}{2}\textbf{x}^T\textbf{A}\textbf{x}+\textbf{b}^Tx+c$, gradient descent with learning rate $\eta$ will diverge if and only if any eigenvalue of $\textbf{A}$ exceeds the threshold $2/\eta$. This bound is sometimes known as the \eos  \cite{cohen2022gradient}, which is conventionally used as an upper bound for $\eta$ to prevent divergence of loss \cite{lecun1992, granziol2020lr}. However, \citet{cohen2022gradient} have shown that despite instabilities, gradient descent can continue to decrease the objective function consistently over long timescales. Alternatively, \citet{lewkowycz2020large} predicted a \textit{catapult} regime of learning rates where gradient descent is unstable, but through instabilities, it eventually gets ``catapulted'' into a region with low sharpness. These observations support using large learning rates that are \textit{unstable} to take larger steps across weight space and find better solutions, challenging existing stability theory which recommends $\eta<2/\lambda_\mathrm{max}$ to guarantee non-divergence of loss. We differentiate the learning rate regimes that fall under or exceed this limit as the \gf\ and the \textit{unstable} regimes respectively. 

\textbf{Outlier-bulk decomposition.} Recent studies on the structure of the Hessian \cite{granziol2020lr, papyan2019measurements, papyan2019spectrum, papyan2020traces, sagun2017eigenvalues} have reported a consistent separation of the \textit{outliers} from the \textit{bulk} of the spectrum. Using \textit{random-matrix theory}, \citet{granziol2020lr} showed that the behaviour of the \textit{bulk} can be viewed as the convergence of a large number of random additive fluctuation matrices, thus justifying $\mathcal{H}_\mathrm{emp} = \mathcal{H}_\mathrm{true} + \epsilon$. On the other hand, \citet{papyan2020traces} utilised the generalised Gauss-Newton decomposition $\mathcal{H}=\mathcal{G}+\mathcal{E}$, where $\mathcal{G}$ is the \textit{Fisher Information Matrix}. Through deflation techniques \cite{papyan2019spectrum}, presented empirical evidence that the spectral \textit{outliers} can be attributed to $\mathcal{G}$ and the \textit{bulk} to $\mathcal{E}$. Additionally, both works present empirical evidence for the existence of spectral outliers as networks are initialised. In section \ref{section:main:phase}, we will refer to a generic \textit{outlier}-\textit{bulk} decomposition of the Hessian inspired by these results:
\begin{align}
    \mathcal{H} &= \mathcal{H}_\mathrm{out} + \mathcal{H}_\mathrm{bulk} = V_{o}W_{o}V_{o}^{T}+\mathcal{H}_\mathrm{bulk}  \label{eqn:hess_decomp}
\end{align}
where $\mathcal{H}_\mathrm{out}$, $\mathcal{H}_\mathrm{bulk}$ are outlier/bulk components and $V_{o}W_{o}V_{o}^T$ the eigen-decomposition of $H_\mathrm{out}$. 

\textbf{Effective parameters.} The number of ``well-determined'' or ``effective'' parameters was used by \citet{djcmThesis} in a Bayesian inference setting to measure the \textit{effective dimensionality} of the model. Using coordinates that condition the Hessian of the \textit{prior} into a unit $n$-sphere, solving for the fixed point of the \textit{log evidence} gives:
\begin{equation}
    \gamma = N_\mathrm{eff} = \sum_{i=1}^{n}\gamma_i = \sum_{i=1}^{n}\left (\frac{\lambda_{b,i}}{\lambda_{b,i}+\alpha} \right )
\end{equation}

where $\lambda_b$s are the eigenvalues of $\textbf{B}$, the objective Hessian. Each $\lambda_{b, i}$ measures how strongly a parameter is determined by the dataset and $\gamma_i \in [0,1]$ measures the strength of the data-determination relative to the \textit{prior} in direction $v_i$. In the \app, we show that each individual ``well-determined'' or ``effective'' eigen-direction of the Hessian controls specific degrees of freedom that correspond to good generalisation, which represent the connection from \neff\ to \EF. 

\textbf{Computation.} The computation and storage of the full Hessian are expensive. We take advantage of existing auto-differentiation libraries \cite{jax2018github} to obtain the Hessian Vector Product (HVP) $\mathcal{H}v = \grad_\theta((\grad_\theta \mathcal{L})v) $, with \citet{pearlmutter}'s trick. We then use the HVP to compute the eigenvector-value pairs, $(\lambda_i,v_i)$, with the Krylov-based Lanczos iteration method \cite{Lanczos:1950zz}. See \app\ for details. 

\section{Main Experiments}\label{section:main}
Our experiments are conducted on fashionMNIST \cite{xiao2017fashionmnist}, a small but challenging benchmark for classification. We provide extensive empirical evidence on CIFAR-10 in Section \ref{section:experiments}.

\subsection{\textbf{S}harpness \textbf{A}djusted \textbf{N}umber of \textbf{E}ffective parameters}\label{section:main:sane}
We introduce \sane, a parameter for model performance that scales the value of $\alpha$ by an eigenvalue $\lambda_\mathrm{scale}$ of the Hessian: \\
\begin{equation}
    \mbox{\sane} = \sum^{n}_{i}
    \left (\frac{\lambda_{i}}{\lambda_{i} + \alpha \lambda_\mathrm{scale}} \right )
\end{equation}
Like \neff, \sane\ leverages the connection to the eigenvectors of the Hessian while being more suitably adapted to the changes in curvature from the instability of gradient descent with large learning rates. Since each eigenvector of the Hessian controls specific degrees of freedom, \sane\ is an attempt at measuring \EF. For our experiments in this work, we use $\lambda_\mathrm{scale}=\lambda_2$, and we encourage the exploration of other choices of $\lambda_\mathrm{scale}$ \footnote{We found preliminary evidence that the effect of progressive sharpening tends to be focused on $\lambda_\mathrm{max}$, so we use an alternative large $\lambda$ (in this case $\lambda_2$) to provide a more stable measure.}. We note that \sane\ does not require significant extra computations compared to \neff. 

\begin{figure}[t]
\centering
\includegraphics[width=1.0\textwidth]{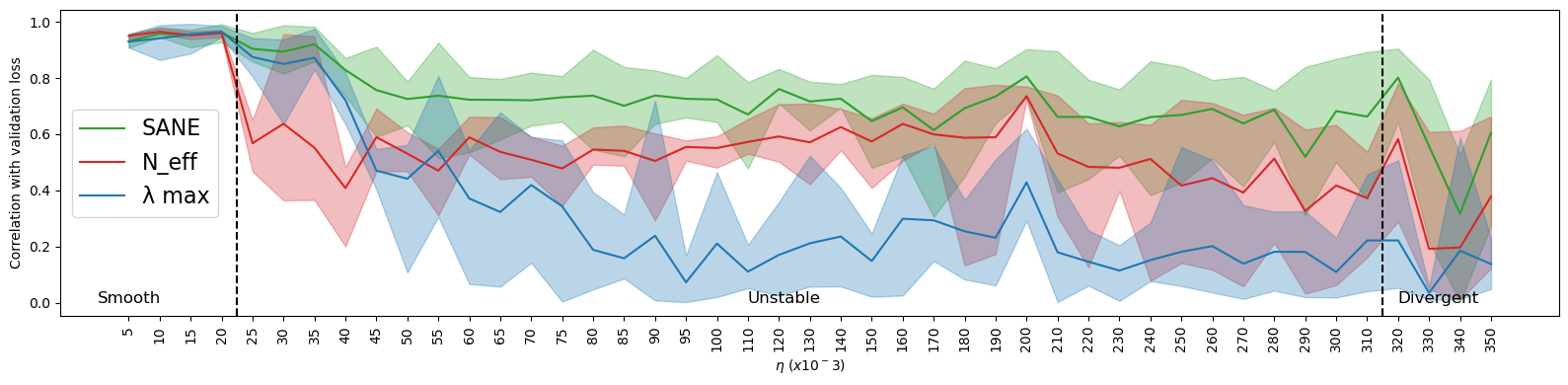}
 \begin{subfigure}[b]{0.32\textwidth}
     \centering
     \includegraphics[height=2cm]{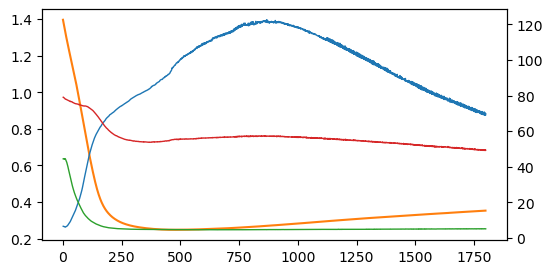}
     \caption{$\eta=0.01$, \gf}
     \label{fig:traj1}
 \end{subfigure}
 \hfill
 \begin{subfigure}[b]{0.32\textwidth}
     \centering
     \includegraphics[height=2cm]{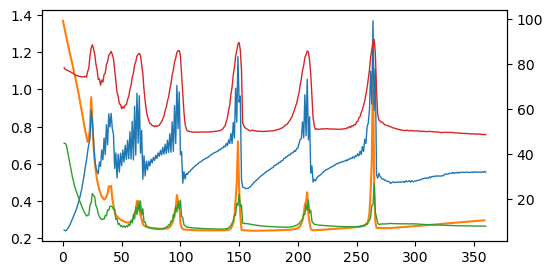}
     \caption{$\eta=0.05$, low instability}
     \label{fig:traj2}
 \end{subfigure}
 \hfill
 \begin{subfigure}[b]{0.32\textwidth}
     \centering
     \includegraphics[height=2cm]{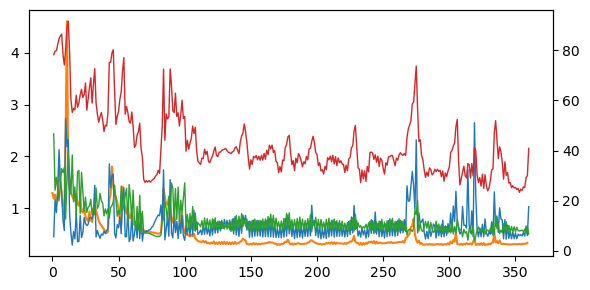}
     \caption{$\eta=0.35$, high instability}
     \label{fig:traj3}
 \end{subfigure}
\caption{\textbf{\sane \ strongly correlates with generalisation along training trajectories up to large $\eta$}. \\ \textbf{Main figure:} correlation with validation loss along trajectories with shaded error bars. The \lr \ regimes are separated by black dotted lines. \textbf{a), b), \& c)}: optimisation trajectories at different \lr s. \textbf{Left axis:} loss; \textbf{Right axis:} \sane, \neff, \& \lammax. While all metrics correlate with validation loss in the \gf\ regime, only \sane\ maintains its performance as learning rates enter unstable regimes. }

\label{fig:correlation}
\end{figure}

\textbf{Generalisation.} We showcase the performance for \textit{post-training} model comparison in Table \ref{tab:corr-post-training}. By design, \sane\ maintains the connection to model performance and generalisation through \textit{effective dimensionality} as the curvature of the loss landscape sharpens over training. This is because, along the training trajectory, the movements of the Hessian spectral outliers are concurrent, aligned, and similar in scale. The performance of \sane\ for model comparison \textit{during-training} is shown by our experiments shown in Fig. \ref{fig:correlation}. In the \gf\ regime, we observe that all measures (\sane, \neff, and \lammax) correlate strongly with validation loss along the model trajectories. However, once we move into the \textit{unstable} regime, the performance of both \neff\ and \lammax\ suffer, while \sane\ maintains its performance even to very-large learning rates. \textbf{\sane, a combination of \lammax\ and \neff, is robust to the unpredictability of large learning-rate training} and we observe that this remains true across architectures, dataset sizes, and MSE Loss. We leave details of these experiments to the \app. 

\textbf{Robustness to spikes.} We visualise the training trajectory of three learning rates in the subplots of Figs. \ref{fig:correlation}. It becomes apparent that the degradation of performance for \neff\ and \lammax\ is a consequence of spiking instabilities inherent to \textit{unstable} gradient descent. In the \gf\ regime, where correlations for \neff\ and \lammax\ with loss are near-optimal, we note that the curves for loss and the measures are simple and continuous. Once we enter the \textit{unstable} regime, instabilities of gradient descent take the form of large \textit{spikes} in loss and measures. \lammax\ oscillates as it approaches the \eos, which is described \cite{cohen2022gradient} as the result of \textit{progressive sharpening} pushing neural weights into sharper regions of the loss landscape while the \eos\ acts as an upper bound, restraining \lammax\ and counteracting \textit{progressive sharpening}. The oscillations uniquely affect \lammax, thereby it weakens the link between curvature and generalisation. While \neff\ is stable between \textit{spikes}, the peak values during instabilities approach the Lanczos iteration limit, which suggests that an insufficient choice of \alf\ has led to a saturation of this measure. On the contrary, \sane\ is relatively stable between spikes and its peaks follow the scale and timing of the spikes well. Since we expect \EF\ measures to be stable in \gf\ learning regimes, the stability of \sane\ in \textit{unstable} regimes offers a suggestion as to why it maintains its relationship to model performance and generalisation.

\textbf{Choice of \alf.} For both \sane\ and \neff, the strength of the parameter \alf\ is a critical scaling factor to the interpretation of the measured \EF. While \citet{djcmThesis} estimates the model evidence to compute an optimal \alf, this approach is not feasible given our interest in model performance during training. More work is required for a direct interpretation of the values of \sane\ and \neff\, so in this work we compare the measures with the \textit{scale-invariant} correlation coefficient. 

\textbf{Over-fitting.} For the \gf\ trajectory in Fig. \ref{fig:traj1} and \textit{unstable} trajectory in Fig. \ref{fig:traj2}, \sane\ remains stable between spikes, while loss increases and \neff\ decreases. These observations reflect over-fitting as validation loss increases despite the stability of \textit{effective dimensionality}. Repeated observations suggest that the connection between \sane\ and generalisation does not address over-fitting. Therefore, in practice, the use of early stopping in conjunction with \sane\, when a robust validation dataset is available, is encouraged. For model comparison \textit{post-training}, we compare separate results for final models (trained to a pre-determined epoch) and early-stopped models. We note the degradation in performance to \sane\ when early-stopping is not included. We leave the study of \sane\ with early stopping, but without the validation set \cite{mahsereci2017early}, as future work.
\begin{table}[t]
    \centering
    \begin{tabular}{ ||c| c c c ||}
    \hline 
     &\sane & \neff & \lammax \\ [0.5ex] 
     \hline
     \hline
     Early-stopped models &\textbf{0.54} & -0.12 & -0.26 \\  
     \hline
     Final models & \textbf{0.42} & -0.15 & -0.18    \\
     \hline
    \end{tabular}
    \vspace{3pt}
    \caption{Correlation of metrics to validation loss. Total 225 models - trajectories plotted in Fig \ref{fig:correlation} }
    \label{tab:corr-post-training}
\end{table}

\subsection{The phases of gradient descent instabilities} \label{section:main:phase} 

In Section \ref{section:background}, we introduced a 3-component ($W_o$, $V_o$, $H_\mathrm{bulk}$) decomposition of the Hessian (Eq. \ref{eqn:hess_decomp}). In Fig. \ref{fig:correlation} we showed the non-monotonic behaviour of gradient descent. The interplay of the Hessian components (\textbf{sharpness} $W_o$, \textbf{rotation} $V_o$, and \textbf{bulk} $H_\mathrm{bulk}$) provide insights into the \textit{spikes} of gradient descent frequently observed. In the following experiments, we compute the gradient misalignment $\overline{S(g)}$ \ and estimate rotational misalignment with $\overline{S(v)} \approx \overline{S(v_\mathrm{max})}$, where $v_\mathrm{max}$ \ is the eigenvector paired with \lammax. Assuming $\mathcal{H}_{out}$ is positive semi-definite and the spectrum of $\mathcal{H}_{bulk}$ has mean zero, we can estimate the scale of $\sigma_{\lambda_\mathrm{bulk}} \propto |\lambda^{-}_{\max}|$, where $\lambda^{-}_{\max}$ is the largest negative eigenvalue. 

\begin{figure}[t]
\centering
\includegraphics[width=1.0\textwidth]{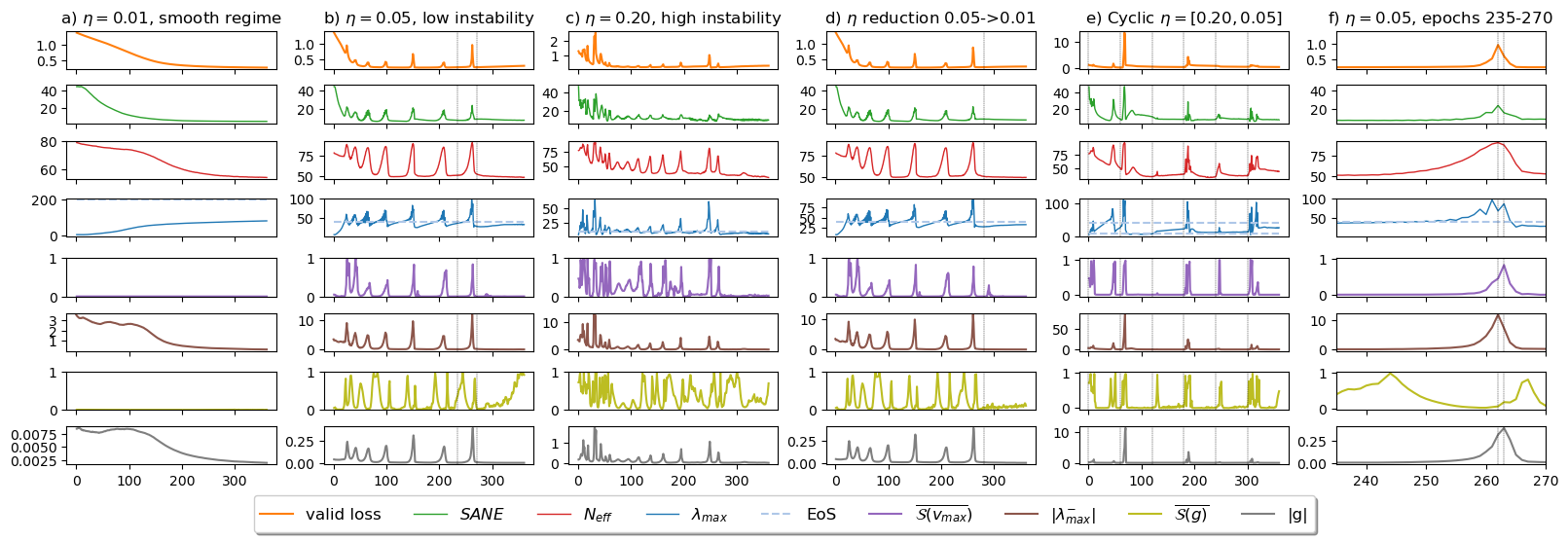}
\caption{\textbf{Instability of gradient descent manifests as peaks in the metrics}. 
\textbf{$\eta$ regimes:} \textbf{a), b), \& c)} constant learning rates; \textbf{d)} $\eta$ reduction ($0.05 \rightarrow 0.01$, epochs $280$-$360$); \textbf{e)} cyclic $\eta$ between $\eta=0.20$ for $10$ epochs and $\eta=0.05$ for $50$ epochs; \textbf{f)} $\eta=0.05$, zoomed in on epochs 235-270. Rows are ordered from top to bottom in the \textbf{legend}. The frequency and magnitude of instabilities grow with $\eta$. }
\label{fig:phases}
\end{figure}

\textbf{Observations.} We visualise various parameters of the Hessian in Fig. \ref{fig:phases}. The \textit{spikes} of gradient descent manifest as peaks in loss, \sane, \lammax, \rotins, \hessvar, and the gradient norm $|g|$. Concentrating on Fig. \ref{fig:phases}f, we see that there exists a \textbf{stable} phase, where \lammax\ oscillates and approaches the \eos. During this time, $|g|$ is low and other metrics (except \gins) are stable. We note the similarity of this phase to the later stages of a \gf\ training trajectory. The building of sharpness toward instability is consistent with the existing theories of stability with gradient descent. However, as the dynamics become increasingly unstable, the trajectory is not prescribed by existing theory. As \lammax\ crosses the \eos, we enter the \textbf{peak} phase, which is accompanied by rises in \lammax, \sane, \rotins, \hessvar, and $|g|$ during this time. Interestingly, the \textbf{peaks of these metrics are not aligned}. Loss, \sane, \neff, and \hessvar\ peak simultaneously at $e_0=261$, which roughly coincides with the centre of the prolonged peak by \lammax. The peak of \gins\ is delayed by one epoch at $e_1=262$, so the largest change in $v_\mathrm{max}$ occurs between $e_0$ and $e_1$. The peak of $|g|$ is also delayed at $e_1$. At the precise moment $e_1$, a new set of solutions are selected and pursued in the \textbf{cooling} phase. The behaviour of \gins is more complex. Specifically, it appears that each \textit{spike} is followed by a peak in \gins, which we believe is evidence to support gradients in \textit{phase transitions} are different from gradients in the \textbf{stable} phase. We believe the peaks in \gins\ preceding the Hessian spike show gradient noise typical to the \gf\ regime and are insignificant since $|g|$ is low. 
\textbf{Phases of instability.} To describe the qualitative behaviour of models, we view non-convex optimisation as travelling through various \textit{loss basins} across the landscape. A loss basin is informally defined as a fixed set of solutions as prescribed by the Hessian, and so changes in the Hessian rotational matrix $V_o$ would signal changes to the loss basin. Our observations capture these behaviours of learning phases: 
\begin{enumerate}
    \item In the \textbf{stable} phase, the model fixates and optimises within a \textit{loss basin}, while \lammax\ grows from \textit{progressive sharpening} without constraint.
    \item As \lammax\ crosses the \eos, the model is forced out of its \textit{loss basin} into a \textbf{peak} where \rotins\ is high, $|g|$ high, and \gins low.
    \item Suddenly, a new \textit{loss basin} is located and pursued, as \rotins, $|g|$ fall and \lammax\ drops below the \eos\ in the \textbf{cooling} phase. 
    \item The post-cooling peak of \gins\ and low $|g|$ signals the start of another \textbf{stable} phase and the cycle repeats. 
\end{enumerate} 

\begin{figure}[t]
\centering
 \begin{subfigure}[b]{0.32\textwidth}
     \centering
     \includegraphics[height=2.4cm]{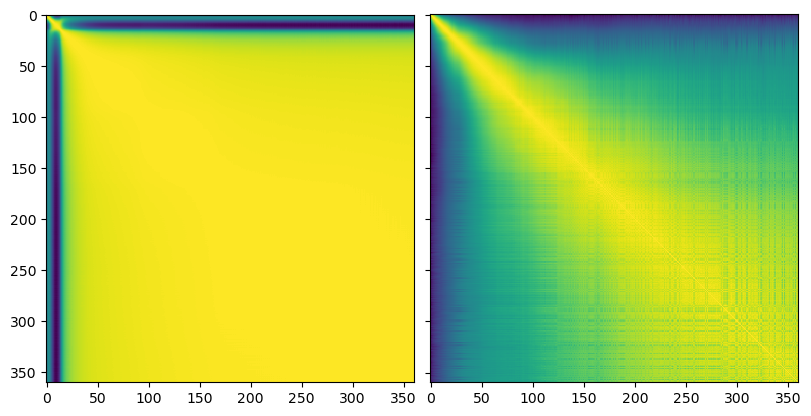}
     \caption{$\eta=0.01$, \gf}
     \label{fig:sim1}
 \end{subfigure}
 \hfill
 \begin{subfigure}[b]{0.32\textwidth}
     \centering
     \includegraphics[height=2.4cm]{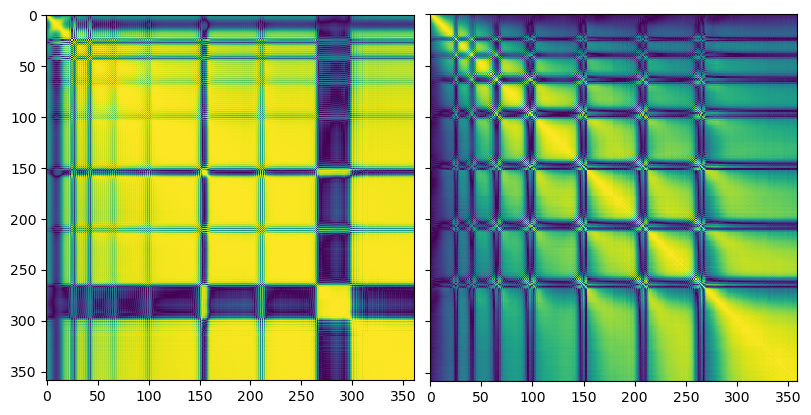}
     \caption{$\eta=0.05$, low instability}
     \label{fig:sim2}
 \end{subfigure}
 \hfill
 \begin{subfigure}[b]{0.32\textwidth}
     \centering
     \includegraphics[height=2.4cm]{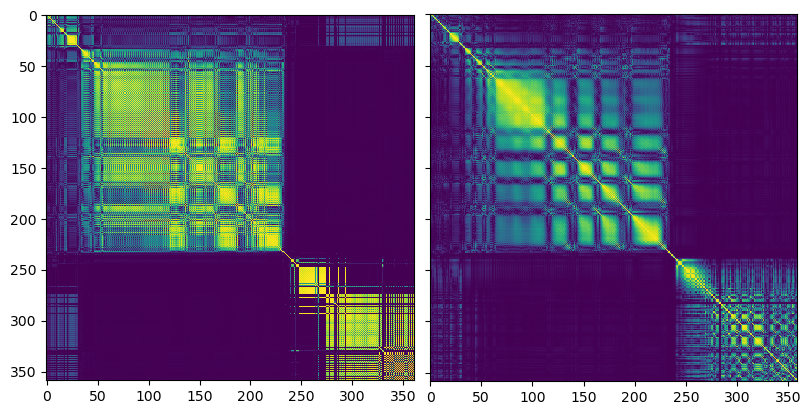}
     \caption{$\eta=0.20$, high instability}
     \label{fig:sim3}
 \end{subfigure}
\centering
\begin{subfigure}[b]{1.0\textwidth}
    \centering
    \includegraphics[width=1.0\textwidth]{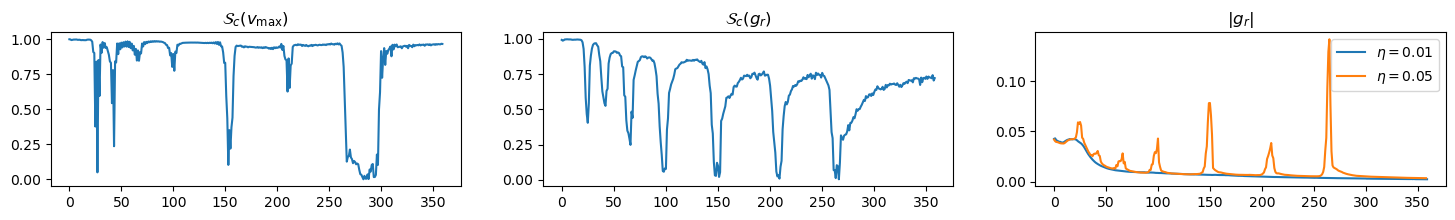}
    \caption{cross trajectory similarity between $\eta=0.01$ and $\eta=0.05$}\label{fig:simAB}
\end{subfigure}
\caption{\textbf{Loss basins between Hessian shifts are similar at low learning rates}. a), b), \& c) pairs: \textit{left miniplot:} $S_c(v_{\mathrm{max},t_i}, v_{\mathrm{max},t_j})$, \textit{right miniplot:} $S_c(g_{r,t_i}, g_{r,t_j})$. \textit{both subplots:} $0\leq t < T_\mathrm{epochs}$. $S_c$s were plotted as a (symmetrical) heatmap using \textit{colourmap viridis}\cite{Hunter:2007}, with deep blue showing zero (no similarity) and yellow showing unity (identical). \textbf{bottom:} $S_c(v_{\mathrm{max},t,\eta_1},v_{\mathrm{max},t,\eta_2})$ and $S_c(g_{r, t, \eta_1},g_{r, t, \eta_2})$ for $\eta_1=0.01$ and $\eta_2=0.05$. The similarity between $\eta=0.01$ and $\eta=0.05$ and the large differences from $\eta=0.20$ (cross-trajectory similarity plots in the \app) support qualitatively different dynamics between regimes of instability. }
\label{fig:sim}
\end{figure}

\textbf{Loss basin shift.} We investigate the nature of \textit{loss basins} as it shifts between phases. We use, $g_r = \frac{1}{2}(g_t+g_{t+1})$, which is a special case of momentum that explicitly evens out the oscillating gradients to describe the components of the gradient orthogonal to the axes of instability. Fig. \ref{fig:sim} shows the similarity of $v_\mathrm{max}$ and of $g_r$ across epochs of three selected learning rates. We use the $\eta=0.01$ \gf\ regime as a benchmark for comparison. After an initial period, \vmax\ maintains extreme similarity, and $g$s are similar if they are close in distance (epochs). Looking at $\eta=0.05$\footnote{The period of low $v_\mathrm{max}$ self-similarity from epochs 262-290 is a consequence of the inaccurate continuity of eigenvalue-vector pairs across time. Detailed plots of the spectrum (shown in the \app) suggest that the top eigenvalue falls beneath the second at epoch 262, but is later brought back above by \textit{progressive sharpening}. }, the spikes of instability interrupt the cadence of the counterfactual \gf path, but a largely analogous structure of similarity emerges when compared to the benchmark \gf\ trajectory. Specifically, \vmax\ exhibits high similarity, barring the initial learning period and \textit{spikes}, and $g$s are similar if they are close in distance. We plot the similarity across trajectories Fig. \ref{fig:simAB}. The combined evidence suggests that only a minor Hessian shift occurs when the \textit{unstable} trajectory experiences instability for $\eta=0.05$. On the other hand, the $\eta=0.20$ trajectory, belonging to the highly unstable regime, displays an extremely low degree of similarity in \vmax\ or $g$ that we see in the \gf\ trajectory. This suggests that a more significant Hessian shift has occurred. Therefore, we observe that instabilities driven by \textbf{larger learning rates decrease the similarity of Hessians} before and after \textit{spikes}. Interestingly, we show in the \app\ that for $\eta=0.05$ and $\eta=0.20$, each successive \textit{spike} gradually moves the optimiser into regions of the weight space with flatter solutions until the maximum sharpness of the solution falls under the \eos\ to enable convergence. While this demonstrates the \textit{flattening} effects of Hessian exploration, we find throughout our work that the quality of solutions does not necessarily improve as the curvature is decreased. 

\textbf{Learning rate schedules.} Using \sane, we study the effects of popular learning rate schedules. Learning rate reduction is a popular schedule used to provide better solutions towards the end of training. As seen in Fig. \ref{fig:phases}d), learning rate reduction allows \lammax\ to exceed the previous \eos\ for a better fit to the solution in the current \textit{loss basin}, effectively moving optimisation into the \gf\ regime provided $\eta$ is low enough. On the other hand, a cyclic learning rate \cite{smith2017cyclical} schedule can be adopted to eliminate the need to find optimal learning rates and schedules. We find that (Fig. \ref{fig:phases}e), with a suitably high $\eta$ upper bound, a cyclic learning rate can encourage the exploration of different Hessian bases at the cost of a reduced likelihood to converge. However, while the solutions slowly get flatter with large $\eta$ (see \app), the performance of solutions (on validation loss) does not necessarily improve. Fortunately, since \sane\ stabilises quickly after the \textbf{cooling} phase, it allows us to determine the quality of \textit{loss basins} quickly. As a result, a \textbf{cyclic learning rate} schedule between a large and a moderate $\eta$ can strike a balance between the \textbf{exploration and exploitation of loss basins}, forming an active (as opposed to black-box) approach to optimisation. We study the exploration of \textit{loss basins} through cyclic learning rate schedules in more detail in the \app.

\section{Related works}\label{section:related}
\textbf{Large learning rates.} Our work builds on a growing body of studies characterising the behaviour of gradient descent at large learning rates as the relationship between large learning rates and generalisation attracts increasing attention. While \citet{cohen2022gradient} showed that \textit{unstable} $\eta$s can decrease consistently over long timescales, we show that there exist $\eta$s that do not cause the model to diverge entirely, but introduce significant rotations to the Hessian where consistent a decrease in the training objective over long timescales is not guaranteed. Like \citet{lewkowycz2020large}, our work prescribes regimes of learning rates where the optimiser is thrown around by the \eos\ and we describe the dynamics of Hessian rotations at both extremes of \textit{unstable} $\eta$s. While they show the Hessian eventually becomes flat, we show (in the \app) that there is a long timescale trend for successive \textit{loss basins} to be flatter, with some randomness at larger $\eta$s. Additionally, \cite{Kaur2022} shows that by varying learning rates, \lammax\ can be arbitrarily altered without necessarily changing the generalisation performance. Specifically, increasing $\eta$ can lead to a drop in test accuracy when batch size is sufficiently large. Their work suggests that this can be remedied if the ratio of $\eta$ to batch size is kept constant, which is consistent with our observations as our study reports deteriorating performance for learning rates in the \textit{highly unstable} regime. 
\textbf{Effective dimensionality} for model comparison has been studied \cite{moody1991}, in a Bayesian setting \cite{Gull1989, djcmThesis, spiegelhalter1998bayesian}. Most recently, \citet{maddox2020rethinking} showed the connection between \neff\ and posterior contraction in Bayesian linear models. Similar to our work, they present empirical evidence for \neff\ as a measure for generalisation, which we echo. Our study of \EF\ is focused on a large learning rate setting for full-batch gradient descent, where \textit{spikes} in the training trajectory destabilise vanilla \neff\ compared to \gf\ learning regimes. We show the robustness of \sane\ to large learning rates and demonstrate a connection between \EF\ and generalisation in this regime. 

\textbf{Phases of gradient descent} were studied by \citet{Li2022} through the lens of sharpness. They proposed a 4-phase division of the gradient descent trajectory based on their analysis (which was rigorously derived for a 2-layer neural net). The \textbf{stable} and \textbf{peak} phases of our study corresponds to phases \textit{I} and \textit{II} in their proposition, while our \textbf{cooling} covers both phases \textit{III} and \textit{IV}. \citet{Li2022} suggested that phase \textit{III} was separated from phase \textit{IV} and in both phases \lammax\ drops. However, they additionally characterise \textit{III} as when $\lambda_\mathrm{max} > 2/\eta$ and the loss continues to peak; and \textit{IV} as when $\lambda_\mathrm{max}<2/\eta$ and the loss drops. While their analysis is sound, our empirical evidence does not support the precise phase demarcation between \textit{III} and \textit{IV}. From our experiments, the rise in loss as we enter the \textbf{peak} phase begins before the peak in \lammax\ was reached and begins to fall while \lammax\ is still above the \eos. We believe our research efforts into the phases of gradient descent are extremely well-aligned and complimentary with \citet{Li2022}, and we leave the reconciliation of our empirical evidence with their theoretical analyses to future work. 

\textbf{Hessian variance.} In our dissection of the phases of learning, we found high agreement between \sane\ and our estimate of \hessvar through $\lambda^{-}_\mathrm{max}$. \cite{granziol2020lr} has defined the Hessian variance by the limiting spectral density of the fluctuation matrix $\epsilon$ through the semicircle law. If $\epsilon$ is uniquely responsible for the \textit{bulk} spectrum, then we can view \hessvar as an estimate of the noise if we treat $\mathcal{H}_\mathrm{out}$ as a signal, $\epsilon=\mathcal{H}-\mathcal{H}_\mathrm{out}$. With this view, \sane\ measures the inverse signal-to-noise ratio of the Hessian. 

\textbf{Compression.} While we attribute the success of \sane\ to the connection to highly specific eigen-directions of the Hessian that correspond to generalisation, some works have focused on the connection to compression \cite{maddox2020rethinking}. We note the intimate relationship between the two approaches. \citet{shwartz-ziv2019representation} derived an information compression bound between the layers of deep neural networks: $I(T_{k+1};T_k|\theta^{*})\leq \frac{1}{2}\sum_{\lambda^{*}_i \in \Lambda^{*}} log(1+\frac{C_{i}}{\lambda^{*}_{i}}) = M$, where $I(T_{k+1};T_k|\theta^{*})$ represent the mutual information between layers $k$ and $k+1$ evaluated at weights $\theta^{*}$, $\Lambda^{*}$ the \textit{informative} eigenvalues of the matrix $\delta \theta^T \delta \theta$, where $\delta \theta$ is modelled as Brownian motion around $\theta^{*}$, and $C$ the norm of $\theta^{*}$. The $\delta \theta$ process is characterised by a low variance in gradients of informative directions, so low $\lambda^{*}$s indicate less informative directions in the gradient. Using these intuitions, we can informally write $\lambda_\mathcal{H}* \lambda^{*} \propto \mathrm{const}$, where $\lambda_\mathcal{H}$ are the eigenvalues of the Hessian. This means that, for each element in the sum, larger values of $\lambda_\mathcal{H}$ increase the contribution to \sane\ and $M$, so when \sane\ grows the mutual information bound is also increased, suggesting a lower level of compression. Likewise, lower values of $\lambda_\mathcal{H}$ suggest that when \sane\ gets a low contribution, $M$ also increases by a low value, suggesting more compression of data. We leave a detailed study of this connection to future work. 

\begin{table}[b]
    \centering
    \begin{tabular}{ ||c| c c c ||}
    \hline 
     &\sane & \neff & \lammax \\ [0.5ex] 
     \hline
     \hline
     Early-stopped, k=1 & \textbf{0.92} & -0.60 & -0.66 \\  
     \hline
     Early-stopped, k=2 &\textbf{0.94} & -0.52 & -0.59 \\  
     \hline
     Early-stopped, k=3 &\textbf{0.92} & -0.67 & -0.61 \\  
     \hline
     \end{tabular}
     \quad
     \begin{tabular}{ ||c| c c c ||}
     \hline 
     &\sane & \neff & \lammax \\ [0.5ex] 
     \hline
     \hline
     Final, k=1 & \textbf{0.88} & \textbf{-0.88} & -0.69    \\
     \hline
     Final, k=2 & \textbf{0.76} & -0.75 & -0.68    \\
     \hline
     Final, k=3 & \textbf{0.87} & -0.86 & -0.71    \\
     \hline
     
    \end{tabular}
    \vspace{3pt}
    \caption{Correlation of metrics to validation loss on CIFAR-10. Total 32 models}
    \label{tab:corr-cifar}
\end{table}

\section{Further Experiments} \label{section:experiments}
Section \ref{section:main} presented our claims with detailed computations on FMNIST. In this section, we verify our claims with experiments on CIFAR-10\cite{krizhevsky2009learning}, a more challenging dataset for image classification. To scale our computation to deep neural nets, we present an approximation to the full Hessian that exploits the natural ordering of neural weights, For details, please refer to the \app.

\textbf{Approximation to the full Hessian. }
We parameterise a prediction function $\hat{y_i} = f_\theta(x_i)$ with a deep neural net with weights $\theta$. Let the network have layers and outputs ($\phi_k$, $T_k$), $k = 1, 2, ... n_\phi$, ordered from the output layer such that $\phi_{k+1}(T_{k+1}, \theta_{k+1}) = T_k$ and $T_{n_\phi}=\textbf{x}, T_0=f_\theta(\textbf{x})$, then $f_\theta(x_i)=\phi_1(\phi_{2}(...(\phi_{n_\phi}(x_i))))$. We consider the reduced objective $\mathcal{L}^\mathrm{r}_{\theta(k)}=\mathbb{E}_{\bm{T}_k} L_{\theta(k)}(\bm{T}_k)$, where $\theta(k)=[\theta_1, \theta_2, ..., \theta_k]$ to compute the reduced Hessian $\mathcal{H}^\mathrm{r}_k=\grad^2_{\theta(k)} \mathcal{L^\mathrm{r}}_{\theta(k)}$. In other words, we approximate the Hessian of the full deep neural network with a Hessian computed on the first $k$ layers closest to the output layer.
Empirically, while this approximation drastically changes the absolute scale of the resulting eigen-spectrum, the relative scales of measures along the training trajectory are accurate (see \app). Recall the discussion in Section \ref{section:main:sane} on a direct interpretation of the numerical values for these measures, we believe \redhess is useful in the study of \sane\ for deeper neural nets. 

\textbf{Experiments on CIFAR-10. }
We tackle the classification problem of CIFAR-10 with an architecture inspired by AlexNet across 32 training configurations (see \app). The results for post-training model comparison are shown in Table \ref{tab:corr-cifar}. We note the strong performance of \sane\ at both \textit{early-stopped} and \textit{final} comparison points, but there is a clear degradation in performance as we remove early stopping. We note that, compared to Table \ref{tab:corr-post-training}, there is a much stronger negative relationship for \neff\ to validation loss. This is unexpected, and the relationship grows stronger for \textit{final} model comparisons. 

\begin{figure}[t] 
\centering
 \begin{subfigure}[b]{1.0\textwidth}
     \centering
     \includegraphics[width=0.9\textwidth]{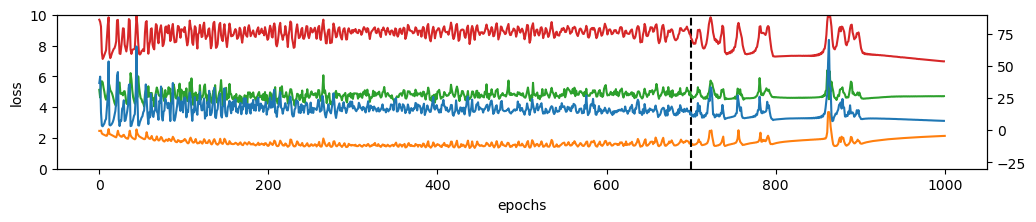}
     \caption{Training trajectory on CIFAR-10, $\eta=0.06$}
        \label{fig:cifartop}
 \end{subfigure}
 \hfill
 \begin{subfigure}[b]{1.0\textwidth}
     \centering
     \includegraphics[width=0.9\textwidth]{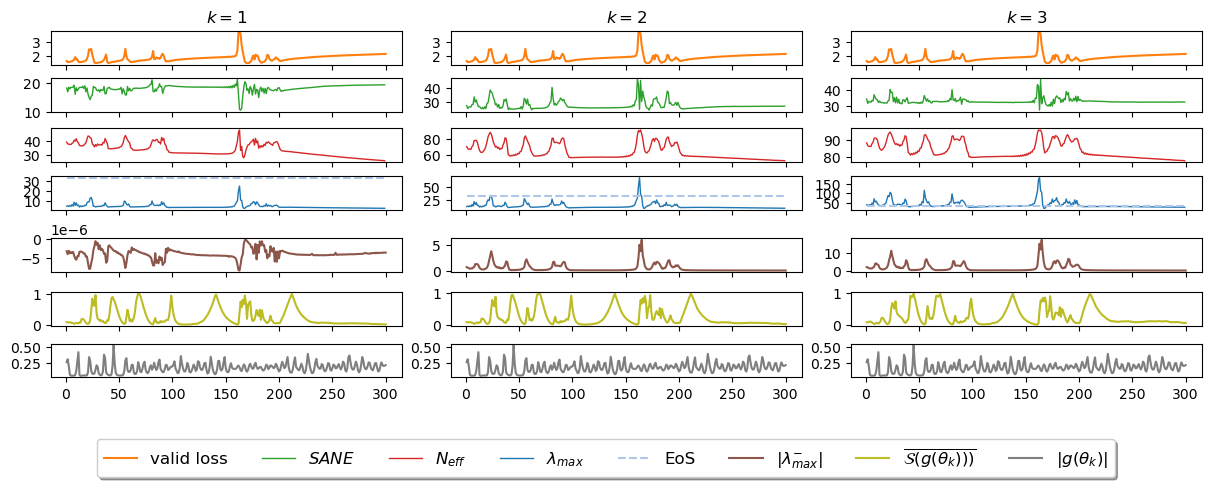}
     \caption{Validation loss and measures; zoomed in to epochs 700-1000}
        \label{fig:cifarbot}
 \end{subfigure}
 
 \caption{\textbf{Training on CIFAR-10 displays phases of instability toward the end}. \textbf{Top:} training trajectory on CIFAR-10 for 1000 epochs. Validation loss (orange and below) plotted on the left axis, while \neff\, \sane\, \lammax\, (in descending order) plotting on the right axis. Axes have been rescaled to promote visibility. \textbf{Bottom:} Plots of validation loss vs measures computed from \redhess\ from epoch $700$, at varying depths $k=1, 2, 3$. We note that the phases of gradient descent are visible from epoch $700$. Additionally, the gradient displays remarkable similarity across the depths of our approximation.
 } \label{fig:cifar}

\end{figure}

Fig. \ref{fig:cifar} shows a specific training trajectory, where we initially observe a large amount of noise. After epoch 700, the trajectory displays the structure of phase transitions, and we visualise the details in Fig. \ref{fig:cifarbot} across depths $k=1,2,3$, where $k=1$ is the output layer only. For each depth, the percentage of model parameters included in the computation was $0.2\%, 9.7\%, 86.5\%$, showing that a more accurate approximation of the full Hessian as k is increased. We note the significant difference in $k=1$ computations vs $k=2,3$, and we note that the output layer is unique in its lack of an activation function. Sadly, we were not able to compute \rotins\ given the limit of GPU memory. We note the clear phases as shown by loss and the measures \sane\, \neff\, and \lammax\, with the corresponding peaks and troughs as described in Section \ref{section:main:phase}. However, the insight of Hessian rotations through \rotins\ and $|g(\theta_k)|$ is not clear, though we note the remarkable similarity of $|g(\theta_k)|$ and \gins\ across depths.

\section{Discussion} \label{section:discussion}
\sane, a novel measure for model performance and generalisation loss, shows promising results in large $\eta$ gradient descent where the curvature of the landscape is famously unstable. However, investigation into gradient descent instabilities has some way to go. The observed positive relationship between validation loss and \lammax, employed by many works to look for flatter solutions, was \emph{not} observed in our experiments. Our work adds to the  body of evidence questioning the relationship between \lammax\ and generalisation and the reconciliation of these observations represents an important research direction. We believe our work presents a step in characterising the phases of gradient descent instabilities. A better taxonomy and understanding of large learning rate regimes could speed up optimisation for practitioners using gradient methods for deep networks. Finally, while we have offered a connection to specific eigen-directions that control the problem-dependent degrees-of-freedom, additional theoretical connections and explanations for why the \EF\ family of methods work could strengthen the connection between model selection and generalisation. We see information compression and Hessian noise play a crucial, but under-explored, role in this connection. 

\newpage

\bibliography{my_bibs}

\newpage

\begin{appendix}

\section{Experimental details}
In this section, we detail the technical details used in the experiments in the main sections.

\textbf{Lanczos iteration.} We can get the HVP using \citet{pearlmutter}'s trick, and we use the \citet{Lanczos:1950zz} algorithm for our Hessian $\mathcal{H}$ computations. Let the number of Lanczos iterations be $n_\mathrm{L}$, the algorithm returns a tridiagonal matrix $T \in \mathcal{R}^{n_\mathrm{L}} \times \mathcal{R}^{n_\mathrm{L}}$. We use $n_\mathrm{L}=100$, and the eigenvalues of the smaller $100 \times 100$ tridiagonal matrix can be readily computed using existing numerical libraries (e.g. \textit{numpy}). The eigenvectors of $\mathcal{H}$ can be computed easily, $V = V_T^T V_\mathrm{L}$, where $V_T$ are the eigenvectors of the tridiagonal matrix $T$ and $V_L$ the Lanczos vectors as secondary outputs from the algorithm. We perform re-orthogonalisation on the matrix of Lanczos vectors after every iteration. Our implementation of jax-powered \cite{jax2018github} Lanczos references a baseline implementation from \textit{https://github.com/google/spectral-density}. 

\textbf{FMNIST.} We train 5-layer MLPs with 32 hidden units in each layer for classification on the FMNIST dataset with cross-entropy loss. Our neural layers use ReLU activation, introduced by \citet{relu}. In Fig. \ref{fig:fmnist-archs}, we train with MSE loss using one-hot encoding for classification. As pointed out by \citet{granziol2020lr}, the batch size $b$ of data can influence the sharpness of the landscape up until a regime of large $b$ where the eigenvalue from $\mathcal{H}_\mathrm{emp}$ dominates the scaling term. Following these intuitions, we compute optimal batch-size $b$ for FMNIST, and found that beyond $b=1000$, the Hessian at initialisation did not reduce significantly in sharpness. This implies that $\mathcal{H}_\mathrm{batch}$ is no longer dominated by the scaling term and so is a $\lambda_{i, b} \approx \lambda_{i, \mathrm{emp}}$. As a result, $b=1000$ represents our full training dataset on FMNIST. To ensure classes are well-represented in the training dataset, we construct the dataset with the first $4$ classes of FMNIST, so that each class will be represented by $\sim 250$ instances in the training dataset. The evaluation set is the same size, $b_\mathrm{eval}=1000$. 

\textbf{Epoch compensation (for FMNIST).} We trained models to $360$ epochs in our experiments plotted in Figs \ref{fig:correlation}, \ref{fig:phases}, \ref{fig:sim}. We observed that models in the \gf\ learning regime required more epochs to travel the same distance in the loss landscape since $\eta$ is small. To ensure an apples-to-apples comparison, we compensated low $\eta$s with additional training epochs $n_e = 360*\eta_b/\eta, \eta <= \eta_b$ using $\eta_b=0.05$ as a marker for the beginning of the \textit{unstable} regime. For $\eta$s larger than $\eta_b$, these trajectories experienced instabilities along the \eos\. For these models, the number of functional evaluations becomes critical toward learning a good solution, so we kept $n_e=360, \eta>=\eta_b$. This is why the subplot in Fig. \ref{fig:traj1} plots $1800=360*0.05/0.01$ epochs. Epoch compensation also affects Fig. \ref{fig:phases} and Fig. \ref{fig:sim}, where for $\eta=0.01$, $360$ epochs are plotted to show a trajectory of $1800$ epochs. 

\textbf{CIFAR-10.} We train modified versions of AlexNet, introduced by \citet{alexnet}, on CIFAR-10 with cross-entropy loss. The network architecture uses $2$ sets of convolution \& max-pool blocks. Convolution layers are structured as ($64$ features, $(5,5)$ kernel, $(2, 2)$ strides) and max-pool as ($(3,3)$ kernel). This structure is followed by $2$ dense (fully-connected) layers with $382$ and $196$ hidden units respectively, and finally an output layer, following the modifications of \citet{keskar2017}. Our model uses ReLU activation \cite{relu}. Similar to FMNIST, we computed an optimal reduced batch size from the full training set, in this case, $b=5000$. All $10$ classes are used in this task, so each class is represented by $\sim 500$ instances in the training dataset. The evaluation set is smaller, $b_\mathrm{eval}=1000$. All models were trained to $1000$ epochs. Unlike FMNIST, no \textit{epoch compensation} was needed for lower $\eta$s since the $\eta$s used were not in the \gf\ regime. 

\textbf{Early-stopping.} To determine early stopping, we traversed validation loss in reverse (backward) order after training to the full number of epochs, and stopped when the next loss is larger than the current loss. This represents a more accurate scheme of early-stopping than typical procedures used \textit{during-training}.

\textbf{Fig. \ref{fig:correlation} \& Table \ref{tab:corr-post-training}.} We plot the correlation between metrics and validation loss of $225$ models ($5$ models each at $45$ learning rates) along the training trajectory in Fig. \ref{fig:correlation} and \textit{post-training} in Table \ref{tab:corr-post-training}. The error bars in Fig. \ref{fig:correlation} were the extremal values from the $5$ models, and we indicate an informal \textbf{Divergent} regime where the learning rate is large enough to observe divergence when training among any of $5$ seeds used. We ignored divergent trajectories for our correlation computations. 

\textbf{Fig \ref{fig:cifar} \& Table \ref{tab:corr-cifar}.} We plot the correlation between metrics and validation loss of $32$ models ($2$ models each at $16$ learning rates) \textit{post-training} in Table \ref{tab:corr-cifar}. A training trajectory with $\eta=0.06$ was shown in Fig. \ref{fig:cifar}. We zoom in to the final $300$ epochs and compute detailed Hessian metrics for each epoch for this stage of training. 

\newpage

\section{Additional synthetic experiments}
\begin{figure}[h] 
 \begin{subfigure}[b]{0.24\textwidth}
     \centering
     \includegraphics[width=\textwidth]{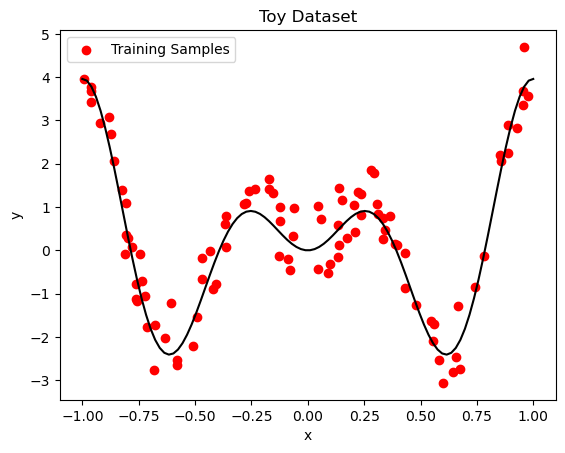}
     \caption{W-reg, train set}
 \end{subfigure}
 \begin{subfigure}[b]{0.24 \textwidth}
     \centering
     \includegraphics[width=\textwidth]{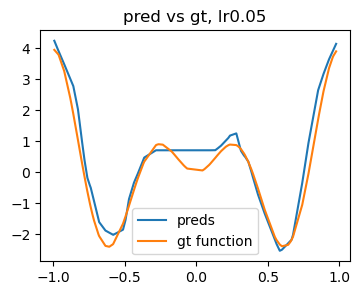}
     \caption{W-reg, sample preds}

 \end{subfigure}
 \begin{subfigure}[b]{0.24 \textwidth}
     \centering
     \includegraphics[width=\textwidth]{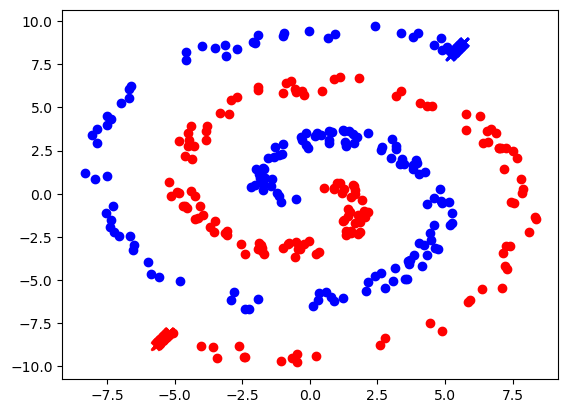}
     \caption{SRC, train set}
 \end{subfigure}
 \begin{subfigure}[b]{0.24 \textwidth}
     \centering
     \includegraphics[width=\textwidth]{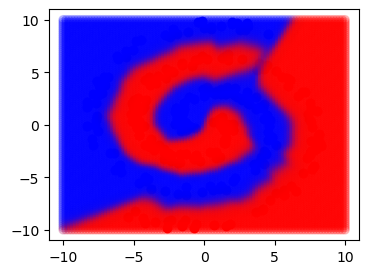}
     \caption{SRC, sample preds}

 \end{subfigure}
 
 \caption{\textbf{Synthetic datasets, \textit{W-reg} \& \textit{SRC}.} Training dataset and sample predictions.} \label{fig:additional-preds}
 
\end{figure}

In this section, we establish the connection of eigenvectors to specific degrees-of-freedom that control performance and generalisation, and we visualise the behaviour of the k-layer approximation to the full Hessian introduced in Section \ref{section:experiments}. Owing to the excessive amounts of compute required for larger datasets and models, we validate these observations on two small synthetic datasets - one for regression and one for classification. Synthetic datasets have the added benefit of allowing a lower dimensional input space to enable more intuitive visualisations of regression predictions and classification boundaries. The regression task fits the function $f(x) = 4x \sin(8x)$, which makes a \textit{W}-shape in the domain $[-1, 1]$, we call this dataset \textbf{W-reg}. For classification, we use a swiss-roll (\textbf{SRC}) dataset which represents a complex transformation from the two-dimensional input space to the feature space. These synthetic datasets are plotted in Fig. \ref{fig:additional-preds}. 

\begin{figure}[p] 
\centering
\makebox[\textwidth][c]{
\begin{subfigure}[b]{1.15\textwidth}
     \includegraphics[width=\textwidth]{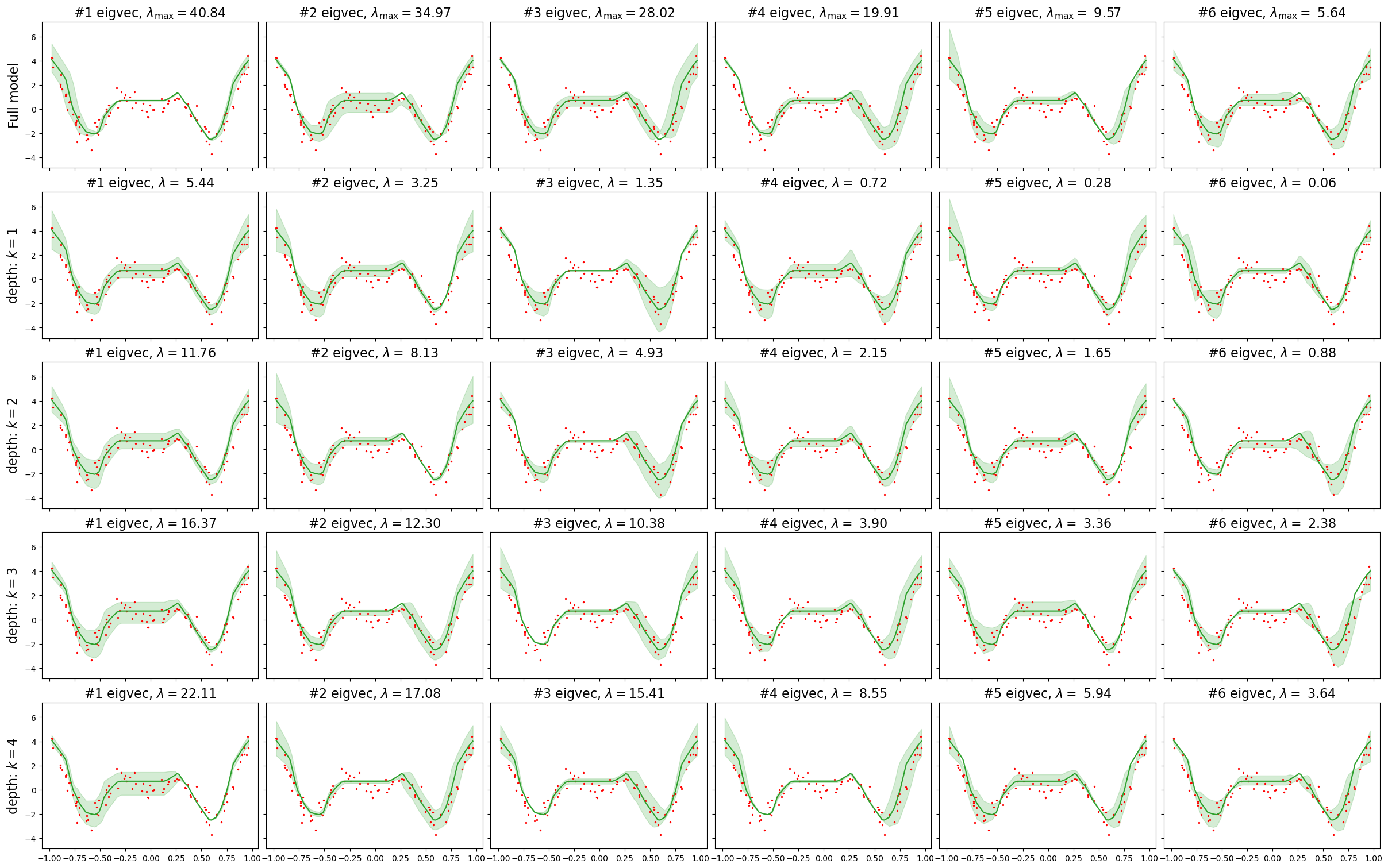}
\end{subfigure}}

 \makebox[\textwidth][c]{
\begin{subfigure}[b]{1.15\textwidth}
     \includegraphics[width=\textwidth]{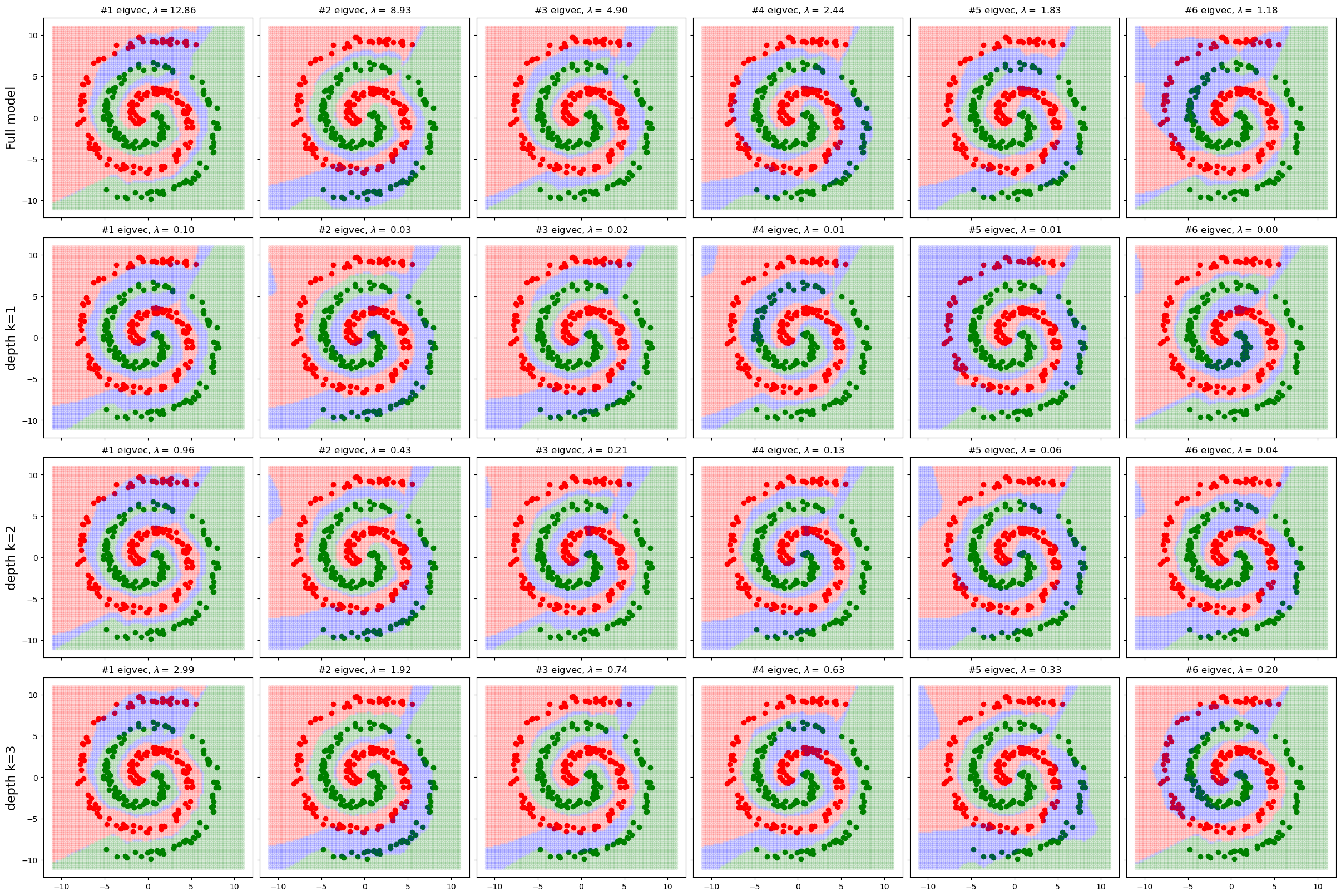}
\end{subfigure}
 }
 
\caption{\textbf{Perturbation studies of top $n=6$ Hessian eigenvectors. } \textbf{Top:} Epoch $2000$ of a $5$-layer MLP trained on \textbf{W-reg}, using $\eta=0.05$, and $k=1, 2, 3, 4$. \textbf{Bottom:} Epoch $2000$ of a $4$-layer MLP trained on \textbf{SRC}, using $\eta=0.05$, and $k=1, 2, 3$. We note that perturbations on eigen-directions of Hessians correspond to uncertainties in local regions of the solution. These eigen-features change minimally when we move the domain of the Hessian from the output layer deeper into the neural net. } \label{fig:regsrc-k}
 
\end{figure}

\subsection{''Sharp'' Eigenvectors correspond to important degrees-of-freedom}
\textbf{Perturbations.} We train MLPs for both W-reg and SRC using MSE and cross-entropy loss at large $\eta$s. Taking models at the final epoch, we visualise the specific degrees-of-freedom (\dofs) corresponding to the top $n=6$ eigenvectors of the Hessian. These \dofs are evaluated through perturbation theory, allowing us to perturbing the model weights along the eigen-directions and measure the effect on output space. The weights are perturbed as $\theta_{p,i} = \theta \pm p_i v_i$, where $p$ is a scaling factor. Perturbations on eigen-directions of the full-model Hessian, as well as with  k-layer approximations of varying depths, are shown in Fig. \ref{fig:regsrc-k}. In a full-batch and large $\eta$ regime, we expect the sharpness of final models to be determined by the unstable dynamics of gradient descent and by the \eos. As $\lambda$s get increasingly sharp, the landscape in the two-dimensional plane defined by $[v, \mathcal{L}]$ gets increasingly sharp, and so we use a square-root scaling factor (based on the quadratic assumption) $p_i=|\lambda_i|^{\frac{1}{2}} c_p$. The positive and negative perturbations form the boundaries of error bars of the visualisations.

\textbf{W regression.} Focusing on the top subplot of the top plot of Fig. \ref{fig:regsrc-k}, we can qualitatively attribute the components of the regression solution to the specific eigenvectors. From the edges, $v_1$ controls the local \dof\ of the left edge of the desired output, and $v3$ the right. Moving inwards, the second bend from both the left and right are controlled by $v_2$ and $v_4$ respectively. $v_2$ also controls the middle plateau, and $v_5$, $v_6$ offer more precise tuning for the sharp turns at the bottom of the \textit{W} shape. We find it surprising that the sharpest eigen-directions appear to control salient \dofs that correspond to performance in the \textit{local} region, and that these sharpest eigenvectors forms a 'sum of local parts' to generate the whole solution. We note that the ordering of $v_1$, $v_2$, and $v_3$ is coincidentally similar to the frequency of datapoints in the training set within the respective regions of the input domain. Since the empirical Hessian driven by the loss from training samples, we conjecture that the relative sharpness of $v$s are determined by the frequency of the corresponding feature in the training set. 

\textbf{Swiss roll classification.} The perturbations plots for SRC are shown in the top subplot of the bottom plot of Fig. \ref{fig:regsrc-k}. The red and green show the 'unperturbed' decision boundaries, while the error bars on the classification boundary due to perturbation have a blue fill. Given the more landscape compared to W-reg, we note that the sharpest eigen-directions of the Hessian correspond to features that are local ($v_1$, $v_2$, $v_6$, and arguably $v_3$). $v_4$ and $v_5$ focus on the boundaries between the swirls - while they looking similar in shape, the regions of uncertainty prescribed by each feature are different and complementary. 

\newpage

\subsection{The k-layer approximation of the Hessian maintains relative scaling}
\begin{figure}[h] 
 \begin{subfigure}[b]{\textwidth}
     \centering
     \includegraphics[width=\textwidth]{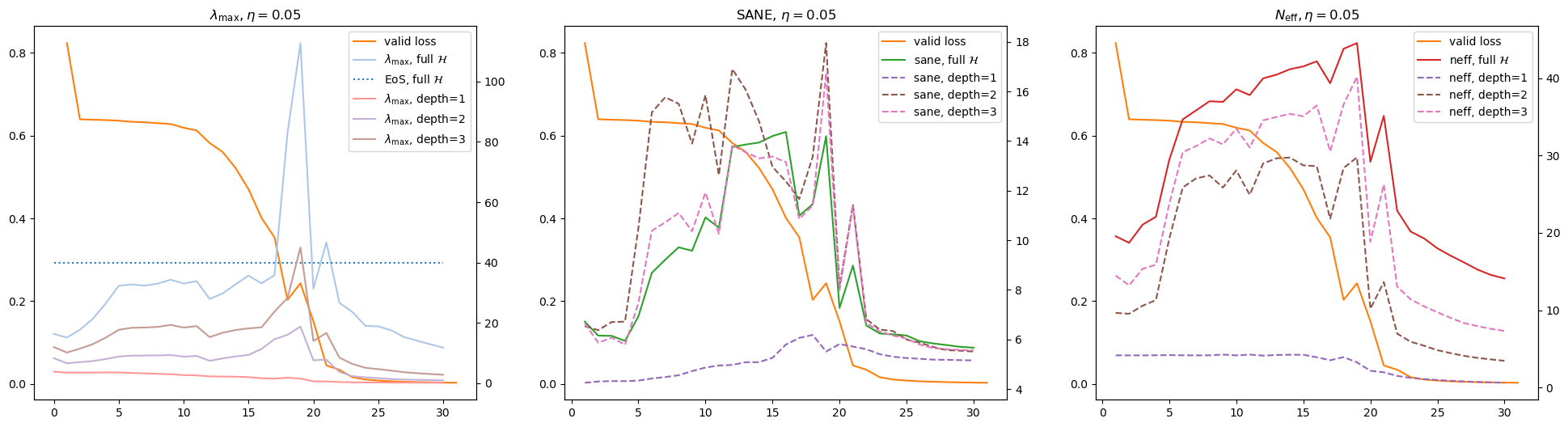}
 \end{subfigure}
 \begin{subfigure}[b]{\textwidth}
     \centering
     \includegraphics[width=\textwidth]{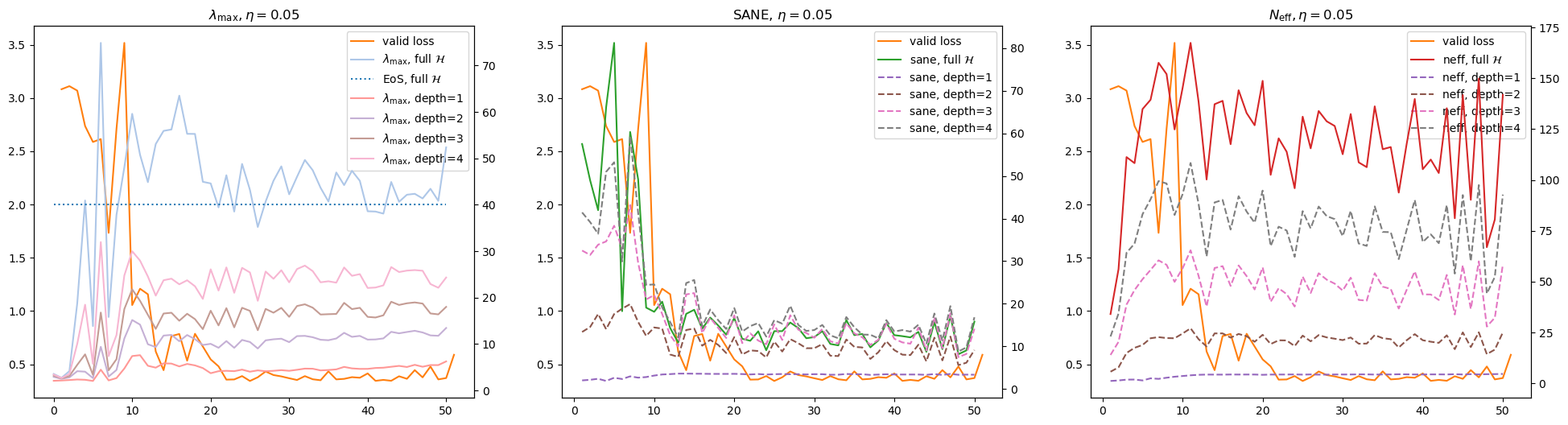}

 \end{subfigure}
 
 \caption{\textbf{\sane, \neff, and \lammax\ computed from the Hessian taken at different depths.} \\ \textbf{Top:} \textit{W-reg}. \textbf{Bottom:} \textit{SRC}. We see that the metrics are increasingly similar in relative scale, and the absolute scales are increasing close to the full model as $k$ is increased. The $k=1$ approximation, which uses only the output-layer, produces flat and uninformative metrics. } \label{fig:metrics-k}
 
\end{figure}

In section \ref{section:experiments}, we introduced the k-layer approximation to the Hessian \redhess\ that exploits the natural ordering of weights. In Fig. \ref{fig:metrics-k} we compare metrics (\sane, \neff, \lammax) computed from \redhess\ to those from $\mathcal{H}$, the full Hessian, on synthetic datasets. In Fig. \ref{fig:regsrc-k}, we perturb the eigenvectors computed from \redhess\ at different depths. From the empirical evidence, we observe that the relative scales of the metrics computed from \redhess\ follows metrics from the full Hessian. This observation was utilised in Section \ref{section:experiments} to provide a connection between metrics from \redhess\ to model performance, since only the relative scales along the trajectory are critical. Secondly, we note that the absolute scale of metrics computed from \redhess\ approaches those from $\mathcal{H}$ as $k$ is increased, which agrees with intuitions. As with Section \ref{section:main}, more work is required for an interpretation of the absolute values of \sane\, \neff, and \lammax. Thirdly, we note that \redhess\ using only the output layer, i.e. $k=1$, computes metrics that are highly uninformative. The low \lammax\ indicate a flat landscape. Despite this, the eigenvectors from \redhess\ $k=1$ correspond to local \dofs that are very similar to that of the full model, and we conjecture that the eigenvectors of the output layer exerts significant control on the specific \dofs for eigenvectors approximated with more layers or from the full model. 

\section{Additional studies}

\subsection{\sane\ is reliable across architectures and loss functions}
\begin{figure}[p] 
\makebox[\textwidth][c]
 {\begin{subfigure}[b]{1.3\textwidth}
     \centering
     
     \includegraphics[width=0.99\textwidth]{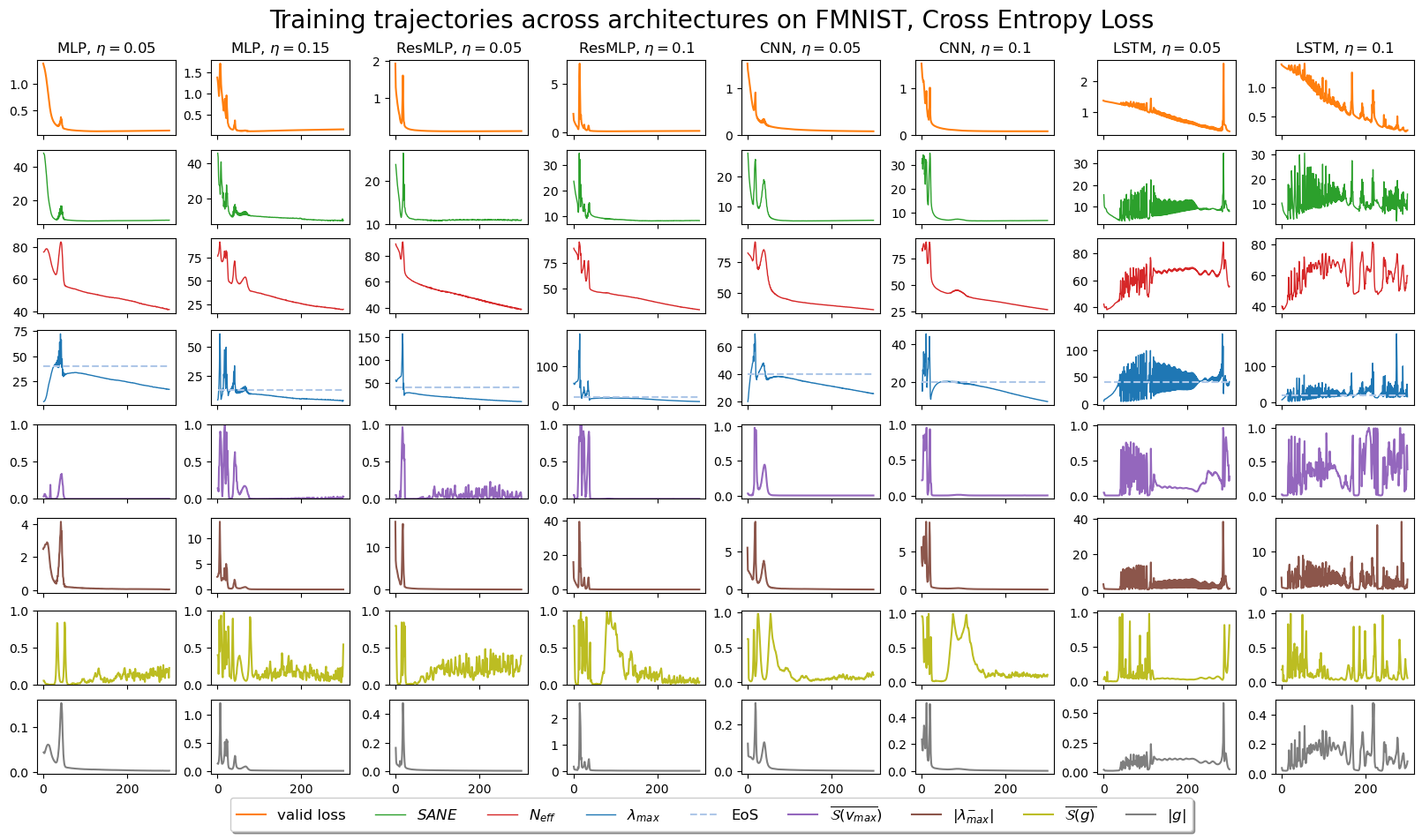}
        \label{fig:fmnist-ce}
 \end{subfigure}}
 \hfill
 \makebox[\textwidth][c]{

 \begin{subfigure}[b]{1.3\textwidth}
     \centering
     \includegraphics[width=0.99\textwidth]{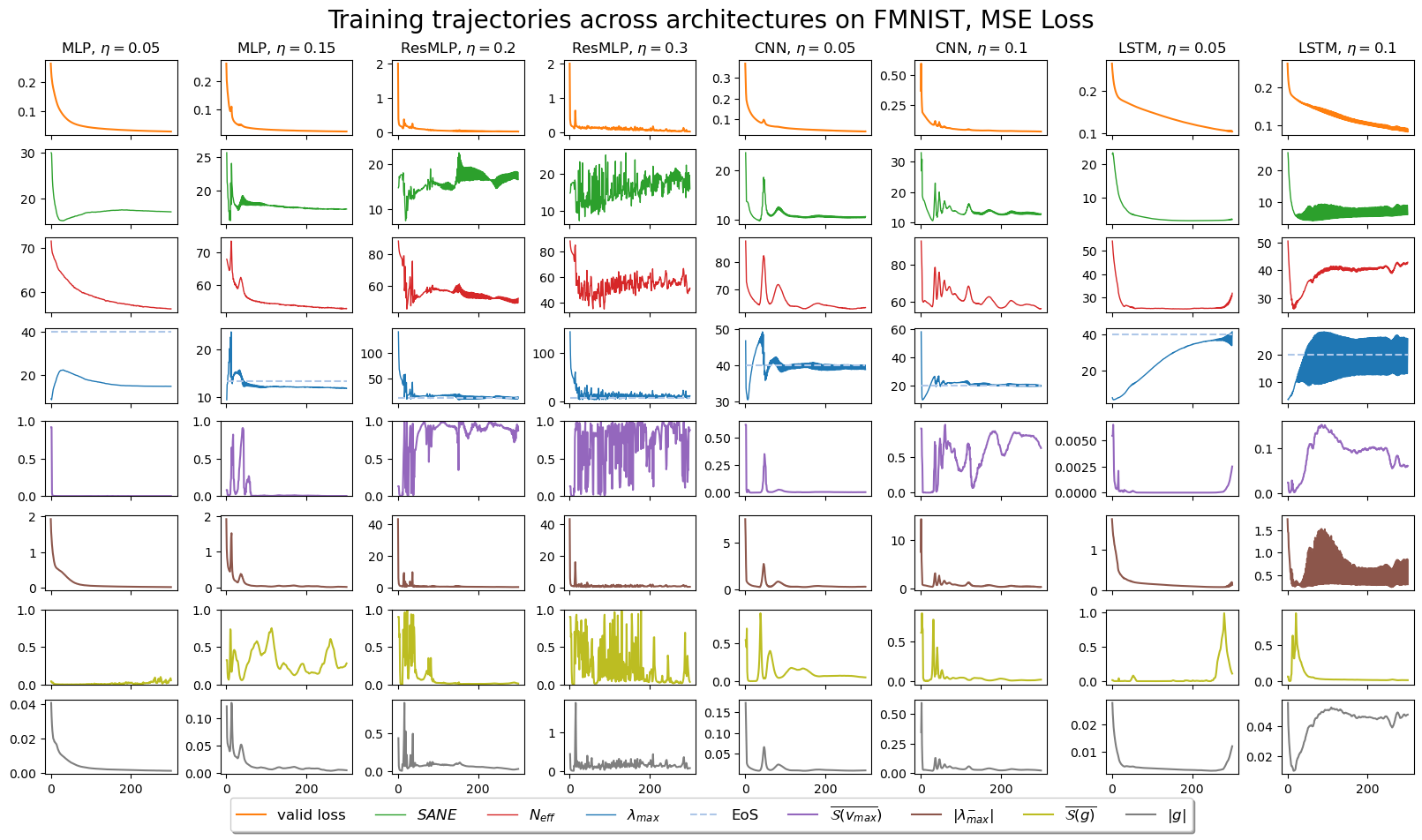}
        \label{fig:fmnist-mse}
 \end{subfigure}}
 
 \caption{\textbf{Instabilities of gradient descent across architectures (MLP, ResMLP, CNN, LSTM) and losses (CE \& MSE).} Unlike \neff\ and \lammax, \sane\ arrives at the 'terminal' value quickly and maintains that value when the Hessian rotation matrix is stable. } \label{fig:fmnist-archs}

\end{figure}

We train an MLP, a ResMLP (MLP with residual/skip connections), a CNN, and an LSTM (\citet{lstm}) on FMNIST with both cross-entropy and MSE loss. The training trajectories are visualised in Fig. \ref{fig:fmnist-archs}. We observe that the phases of instability are observed across architectures and objectives, as well as evidence for \sane\ being a more reliable tool for model comparison, and it tracks the instabilities well and remains constant when the Hessian rotation matrix is constant. Interestingly, the 'tailing-off' behaviour of \lammax\ when the model over-fits in cross-entropy loss, toward the end of training, is not observed when we use one-hot encoded MSE-loss. This observation was made by \citet{cohen2022gradient} which we echo. Despite this, \sane\ still appears more stable than \neff\ in an MSE setting as it arrives at and maintains a constant value. These experiments are an initial  foray into the multitude of architectures, tasks, objectives, and other factors, and we leave the extensive survey of our work as future work.

\newpage

\subsection{Appendix to Fig. \ref{fig:sim}}
\begin{figure}[h] 
 \begin{subfigure}[b]{\textwidth}
     \centering
     \includegraphics[width=\textwidth]{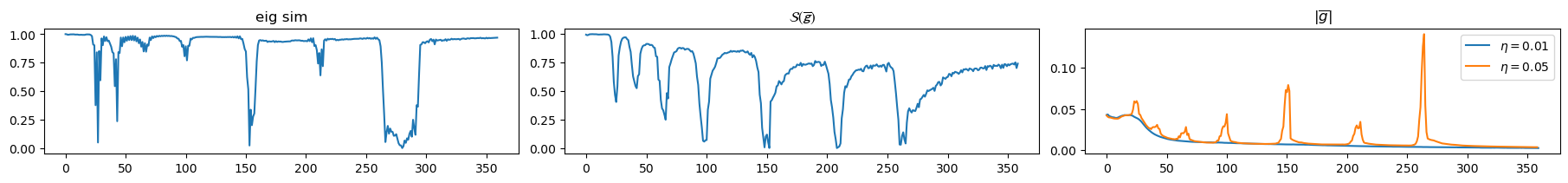}
 \end{subfigure}
 \begin{subfigure}[b]{\textwidth}
     \centering
     \includegraphics[width=\textwidth]{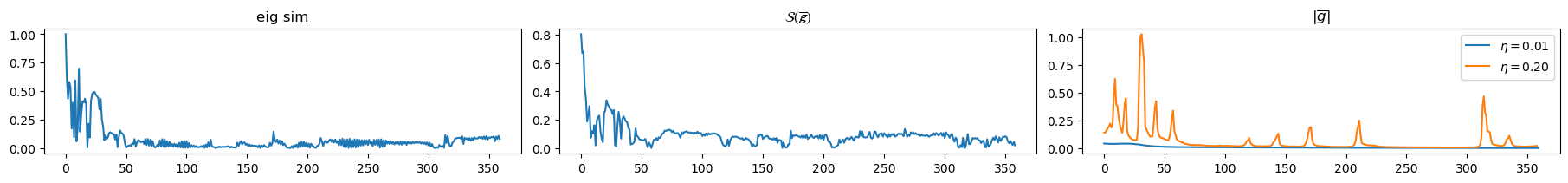}

 \end{subfigure}
 \begin{subfigure}[b]{\textwidth}
     \centering
     \includegraphics[width=\textwidth]{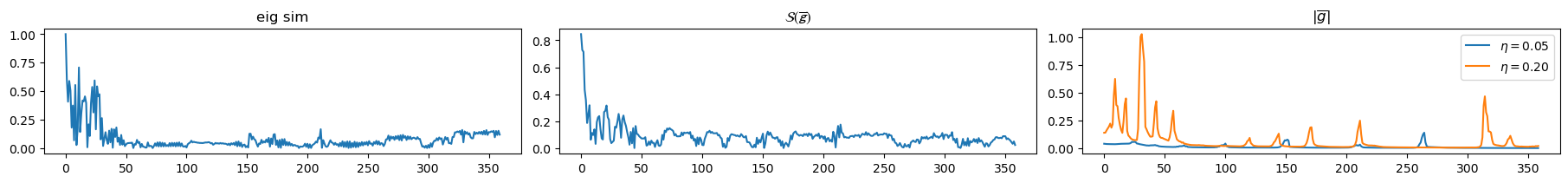}

 \end{subfigure}
 \hfill

 \caption{\textbf{Cross-similarity plots to supplement Fig. \ref{fig:simAB}.} We observe the difference the Hessians of $\eta=0.20$ has to other trajectories, $\eta=0.01$ and $\eta=0.05$. } \label{fig:cross-traj}
 
\end{figure}

\begin{figure}[h]
 \centering
 \   \includegraphics[width=0.7\textwidth]{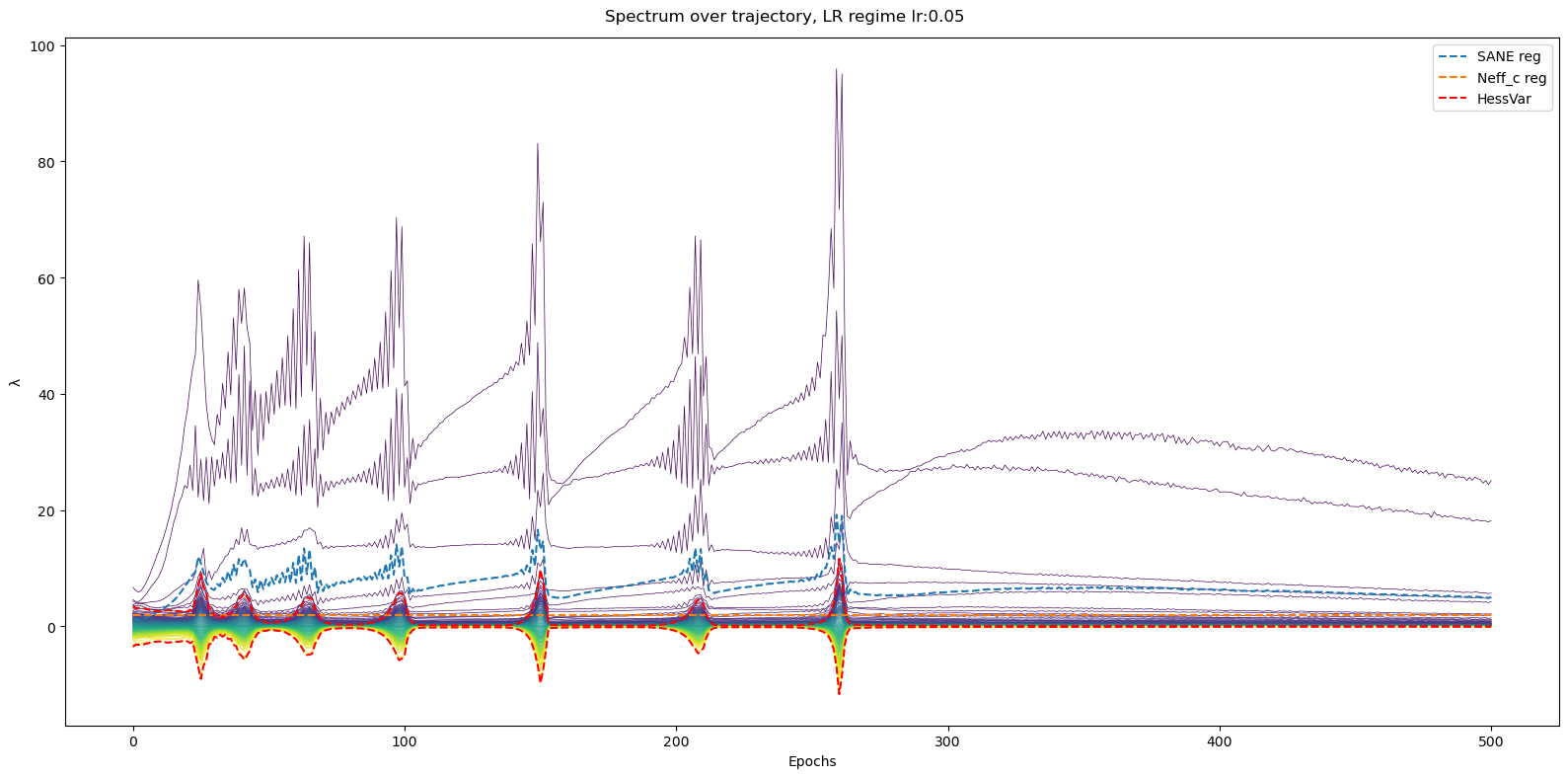}
 \caption{\textbf{Hessian spectrum of the $\eta=0.05$ trajectory. } The limits of \hessvar and the regularising factors of \sane\ and \neff\ are computed along the trajectory. The colormap 'viridis'\cite{Hunter:2007} is used to order the eigenvalues. We observe \hessvar cutting through the constant-valued regularising factor of \neff, which leads to the near Lanczos-limit values of \neff\ reported in the main paper. Additionally, we note the reordering of eigenvalue rankings at epoch $\approx 290$, which suggests the uniqueness of progressive sharpening and validate the gap in similarity in Fig. \ref{fig:sim2}.} \label{fig:spectrum005}

 \hfill
\end{figure}

In Fig. \ref{fig:cross-traj}, we complete Fig. \ref{fig:sim} by plotting the additional cross-similarity figures for $\eta$ pairs: $\eta_1=0.01$, $\eta_2=0.20$ and $\eta_1=0.05$, $\eta_2=0.20$. This supports the claim, made in the main paper, that $\eta=0.20$ experiences significant shifts in the Hessian rotation matrix, while the $\eta=0.01$ and $\eta=0.05$ trajectories are very similar to one another. Additionally, we detail the Hessian spectrum for $\eta=0.05$ in Fig. \ref{fig:spectrum005} to validate the gap between epochs 262-290 in Fig. \ref{fig:sim2}. 

\subsection{Instabilities move the optimiser into gradually flatter regions} \label{sm:flatter}
\begin{figure}[p] 
 \hfill
 
\makebox[\textwidth][c]{

 \begin{subfigure}[b]{1.4\textwidth}
     \centering
     \includegraphics[width=0.99\textwidth]{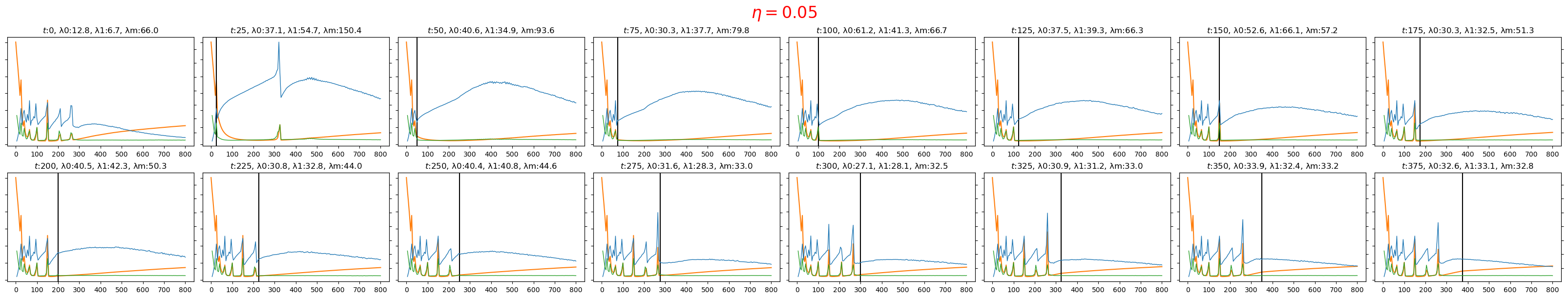}
 \end{subfigure}}
 \hfill
  \makebox[\textwidth][c]{

 \begin{subfigure}[b]{1.4\textwidth}
     \centering
     \includegraphics[width=0.99\textwidth]{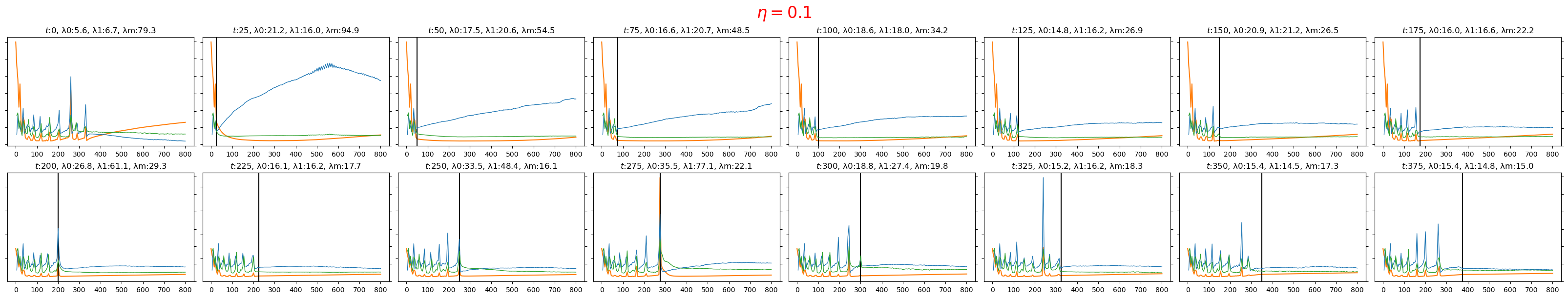}
 \end{subfigure}}
  \hfill
   \makebox[\textwidth][c]{

 \begin{subfigure}[b]{1.4\textwidth}
     \centering
     \includegraphics[width=0.99\textwidth]{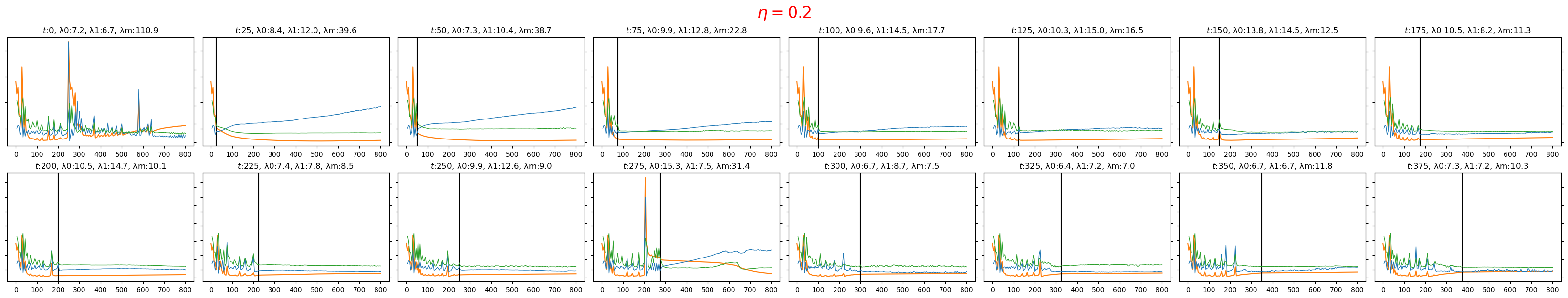}
 \end{subfigure}}
  \hfill
   \makebox[\textwidth][c]{

 \begin{subfigure}[b]{1.4\textwidth}
     \centering
     \includegraphics[width=0.99\textwidth]{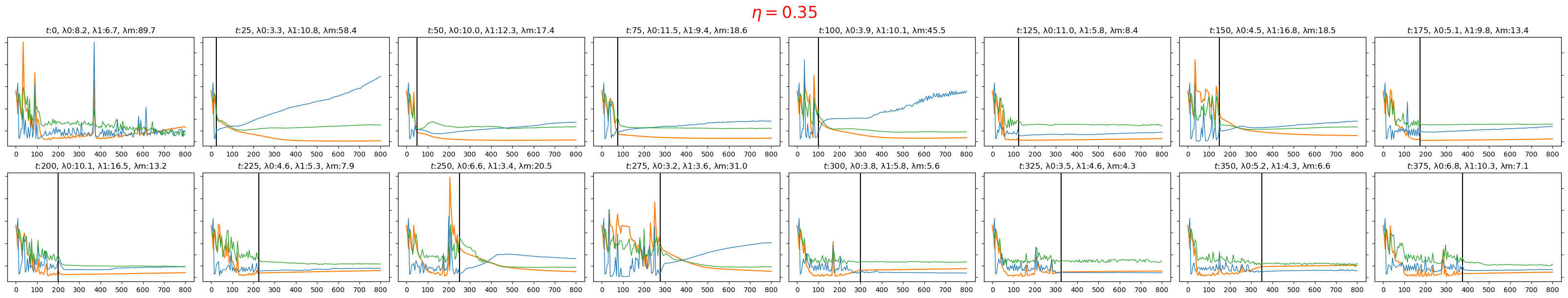}
 \end{subfigure}}
  \hfill
   \makebox[\textwidth][c]{

 \begin{subfigure}[b]{1\textwidth}
     \centering
     \includegraphics[width=0.99\textwidth]{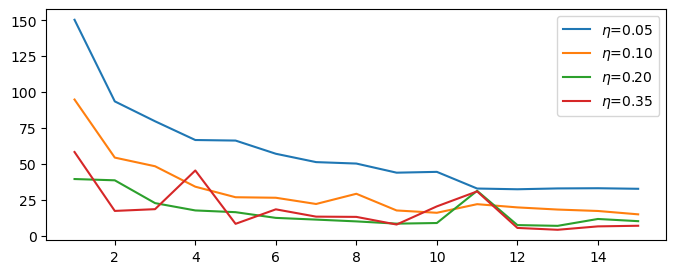}
 \end{subfigure}}

 \caption{\textbf{Delaying $\eta$ reduction allows optimisers to move into flatter solutions. Final \lammax\ of solutions tend to decrease as $\eta$ reduction is delayed.} Varing initial learning rates $\eta_0$ are titled in red. \textbf{Bottom:} summary of final \lammax\ as $\eta$ reduction is delayed. We see a general downward trend that is weakened as $\eta$ is increased. } \label{fig:break}
 
\end{figure}

We support the claim made in Section \ref{section:main:phase} that we found the sharpness to gradually decrease along the training trajectory despite instabilities. This sharpness is defined as the final \lammax\ given a stable Hessian rotation matrix. To evaluate this, we perform $\eta$ reduction to $\eta=0.02$ at different epochs along the training trajectory. $\eta$ reduction has the effect of raising the \eos\ limit, allowing solutions to \textit{progressively sharpen} to their potential. The resulting trajectories are plotting in Fig. \ref{fig:break}, with a summary on the bottom subplot. \textbf{We observe a general trend for maximal \lammax\ of solutions to decrease as we delay $\eta$ reduction.} However, this relationship becomes less clear as the initial learning rate is increased, when the \eos\ is already very limiting. We conjecture that the weakened relationship is a consequence of larger shifts in the Hessian rotation matrix as a result of using large $\eta$s, the latter observation is made at other points of this work.

\newpage

\subsection{Similarity of Hessians through $\eta$ reduction}
\begin{figure}[h] 
 \begin{subfigure}[b]{0.2\textwidth}
     \centering
     \includegraphics[width=0.99\textwidth]{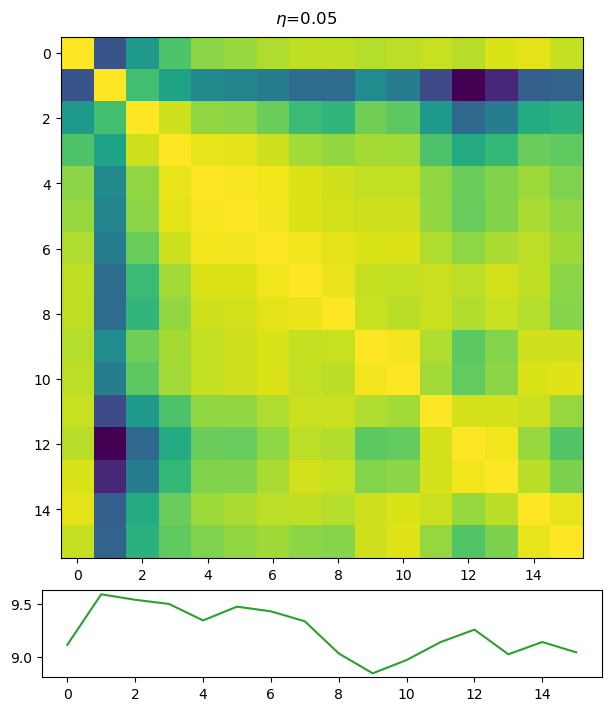}
 \end{subfigure}
 \begin{subfigure}[b]{0.2\textwidth}
     \centering
     \includegraphics[width=0.99\textwidth]{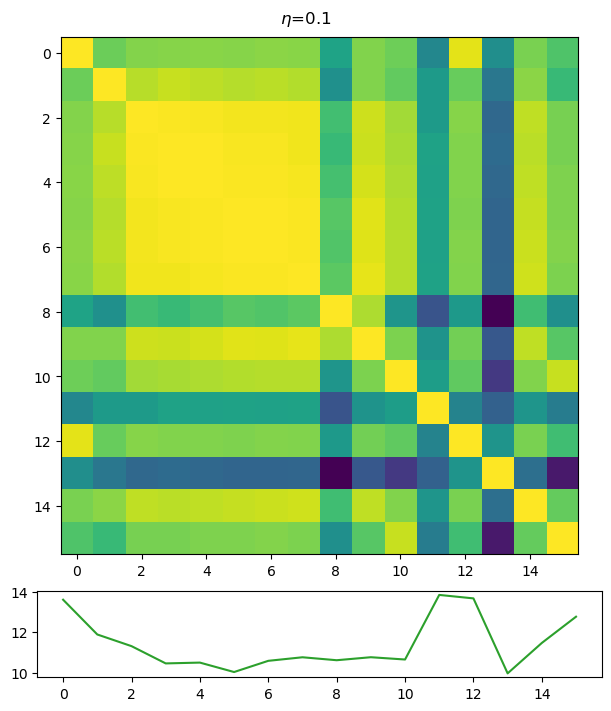}
 \end{subfigure}
 \begin{subfigure}[b]{0.2\textwidth}
     \centering
     \includegraphics[width=0.99\textwidth]{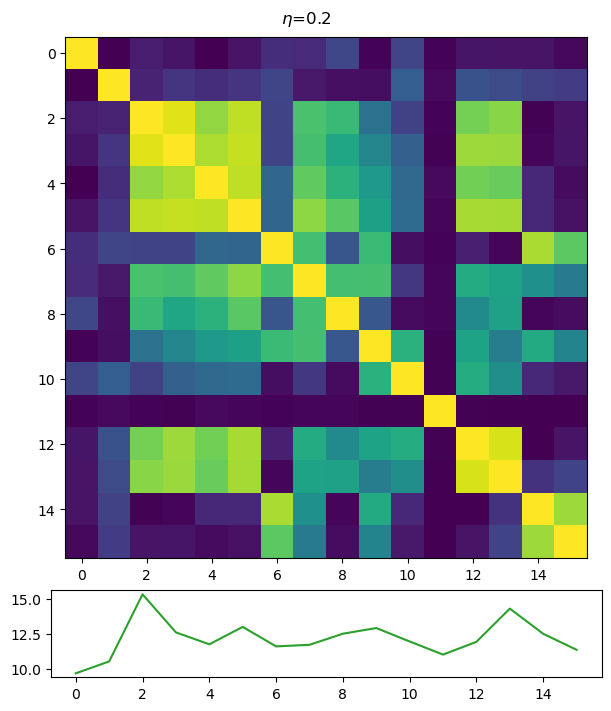}
 \end{subfigure}
 \begin{subfigure}[b]{0.2\textwidth}
     \centering
     \includegraphics[width=0.99\textwidth]{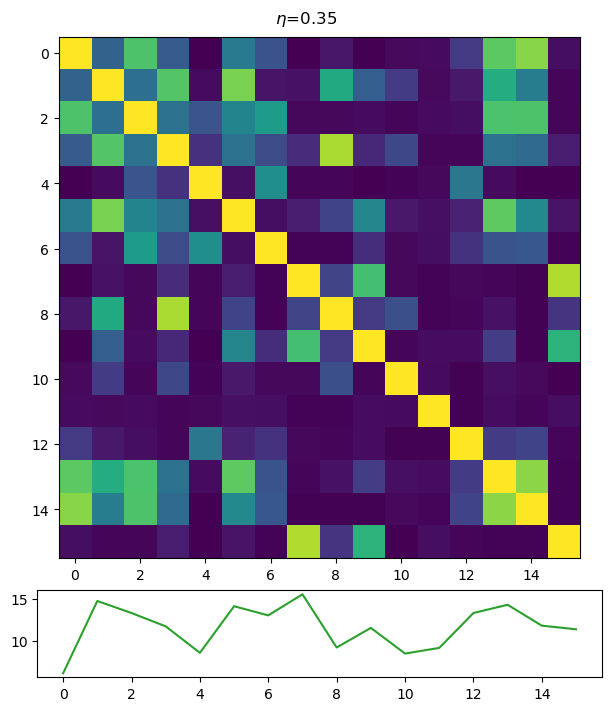}
 \end{subfigure}

\caption{\textbf{Lower learning rates do not shift the Hessian rotation matrix.} \textbf{Top:} $S_c(v_{\mathrm{max},t_\mathrm{final}}, v_{\mathrm{max},t_\mathrm{final}})$ for four learning rates with $\eta$ reduction conducted at different points of the trajectory. All in the $\eta$s are in \textit{unstable} regime. \textbf{Bottom:} Final \sane\ metrics. We observe that lower $\eta$s maintain higher similarity in $v_max$ as the model is trained to completion. This supports our hypothesis that larger $\eta$s shifts the Hessian in a qualitatively different way to low $\eta$s.} \label{fig:breaksim}
 
\end{figure}

We study the similarity of $v_\mathrm{max}$ as we delay $\eta$ reduction. While it is established in Section \ref{sm:flatter} that \lammax\ is flatter as we delay $\eta$ reduction, it is surprising that $v_\mathrm{max}$, taken at the final training epoch, maintains great similarities to other models which are $\eta$ reduced earlier (which have sharper solutions). \textbf{This suggests that for low $\eta$, the optimiser moves solutions into flatter regions with a similar Hessian rotation matrix.} This observation is made from the evidence presented in Fig. \ref{fig:breaksim}, and we observe that larger $\eta$s lead to models that are less well-aligned in $v_\mathrm{max}$.

\subsection{Hessian shifts with cyclic learning rates}
\begin{figure}[h] 
\centering
 \begin{subfigure}[l]{0.2\textwidth}
     \centering
     \includegraphics[width=0.99\textwidth]{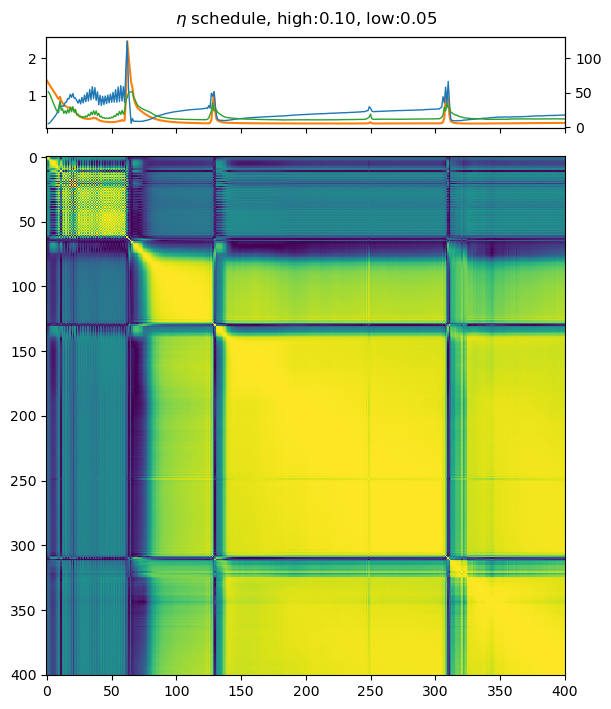}
 \end{subfigure}
 \begin{subfigure}[c]{0.2\textwidth}
     \centering
     \includegraphics[width=0.99\textwidth]{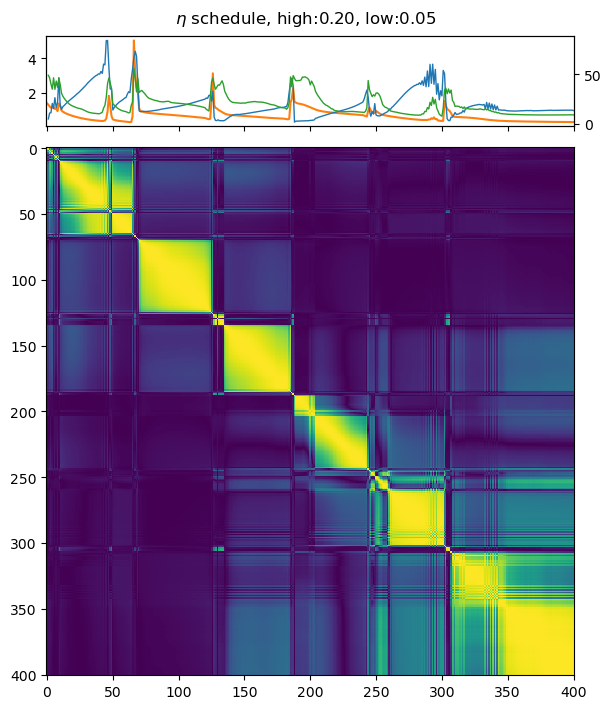}
 \end{subfigure}
 \begin{subfigure}[r]{0.2\textwidth}
     \centering
     \includegraphics[width=0.99\textwidth]{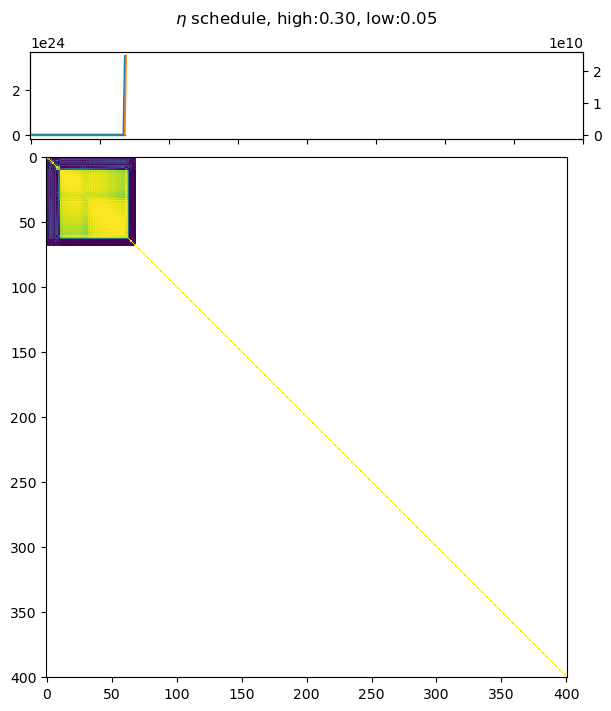}
 \end{subfigure}

\centering
 \begin{subfigure}[l]{0.2\textwidth}
     \centering
     \includegraphics[width=0.99\textwidth]{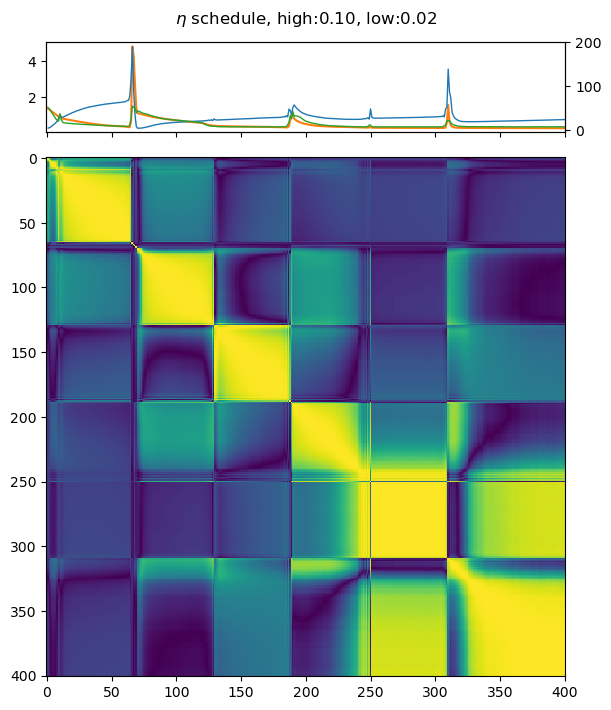}
 \end{subfigure}
 \begin{subfigure}[c]{0.2\textwidth}
     \centering
     \includegraphics[width=0.99\textwidth]{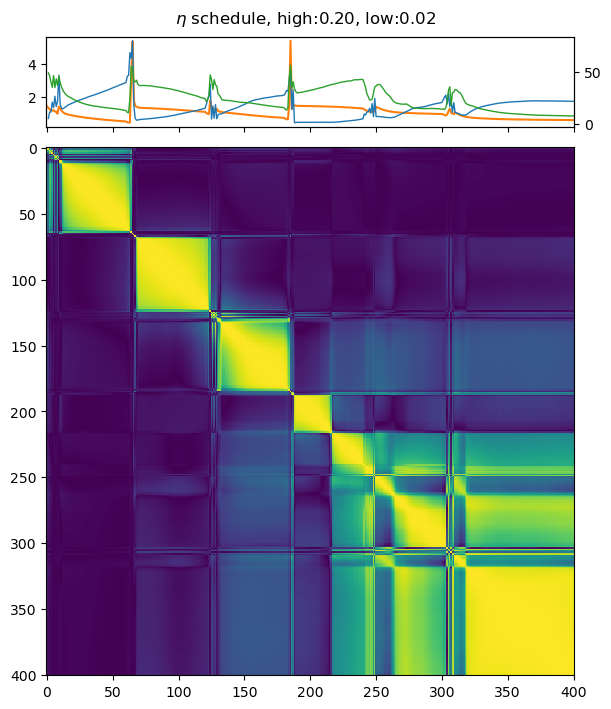}
 \end{subfigure}
 \begin{subfigure}[r]{0.2\textwidth}
     \centering
     \includegraphics[width=0.99\textwidth]{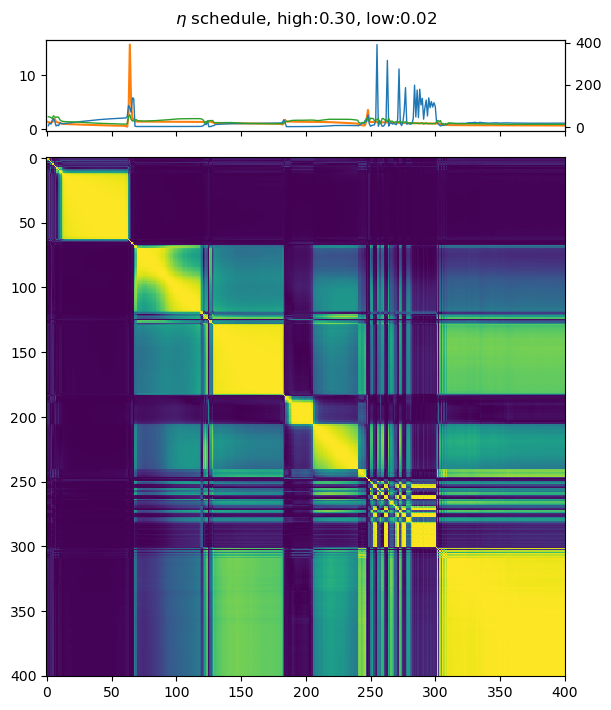}
 \end{subfigure}
  \hfill
 \caption{\textbf{Training trajectories and similarities of six cyclic $\eta$ schedules.} \textbf{Columns:} $\eta_+ \in (0.10, 0.20, 0.30)$. \textbf{Rows:} $\eta_- \in (0.05, 0.02)$. \textbf{Top of subplots:} learning trajectory (validation loss, \sane\, \neff\, \lammax\ plotted). \textbf{Bottom of subplots:} $S_c(v_{\mathrm{max},t_i}, v_{\mathrm{max},t_j})$ of cyclic $\eta$ schedules. The schedule of $\eta_\mathrm{high}=0.30$ and $\eta_\mathrm{low}=0.05$ has diverged during training. \textbf{Cyclic $\eta$ schedules can balance the exploration and exploitation trade-off of \textit{loss basins} depending on selected $\eta$s.} } \label{fig:cyclic}
 
\end{figure}

We study the exploration of \textit{loss basins} under six cyclic $\eta$ schedules through the similarity of $v_\mathrm{max}$. We take $\eta_+ \in (0.10,0.20,0.30)$, situated well within the \textit{unstable} regime, as the upper limits of our schedule; $\eta_- \in (0.02, 0.05)$ are used as the lower limits. $\eta=0.02$ is in the \gf\ earning regime while $\eta=0.05$ is \textit{unstable}. These cyclic $\eta$ schemes use $\eta_+$ for $10$ epochs, before switching to $\eta_-$ for $50$ epochs and repeating. The final $40$ epochs use $\eta_-$. The results are visualised in Fig. \ref{fig:cyclic}, and we see that difference choices of $\eta_+$ and $\eta_-$ can lead to different movements of the Hessian. We observe that large $\eta_+$s encourages Hessian shifts. Surprisingly, it is not the case that using low $\eta_+$ and $\eta_-$ will lead to stagnation in similar \textit{loss basins}. The ratio $\frac{\eta_+}{\eta_-}$ appears to play an important role.

\end{appendix}

\end{document}